\documentclass[sigconf]{acmart}

\usepackage{amsmath,amsfonts}
\usepackage[noend]{algorithmic}
\usepackage{algorithm}
\usepackage{array}
\usepackage[caption=false,font=normalsize,labelfont=sf,textfont=sf]{subfig}
\usepackage{textcomp}
\usepackage{stfloats}
\usepackage{verbatim}
\usepackage{graphicx}
\usepackage{xcolor}
\usepackage{microtype}
\usepackage{graphicx}
\usepackage{booktabs}
\usepackage{mathtools}
\usepackage{pifont}
\usepackage{amsthm}
\usepackage{paralist}
\usepackage{mathrsfs}
\usepackage{newfloat}
\usepackage{listings}
\usepackage[capitalize,noabbrev]{cleveref}
\usepackage{multirow}
\usepackage{colortbl}
\newcommand{\rmnum}[1]{\romannumeral #1}

\copyrightyear{2025}
\acmYear{2025}
\setcopyright{acmlicensed}
\acmConference[CCS '25]{Proceedings of the 2025 ACM SIGSAC Conference on Computer and Communications Security}{October 13--17, 2025}{Taipei, Taiwan}
\acmBooktitle{Proceedings of the 2025 ACM SIGSAC Conference on Computer and Communications Security (CCS '25), October 13--17, 2025, Taipei, Taiwan}
\acmDOI{10.1145/3719027.3744883}
\acmISBN{979-8-4007-1525-9/2025/10}

\begin{document}

\title{FilterFL: Knowledge Filtering-based Data-Free Backdoor Defense for Federated Learning}

\author{Yanxin Yang}
\affiliation{%
  \institution{East China Normal University, Shanghai, China}
  \country{}
}
\email{52275902023@stu.ecnu.edu.cn}
\author{Ming Hu}
\authornote{Ming Hu and Mingsong Chen are the corresponding authors.}
\affiliation{%
  \institution{Singapore Management University, Singapore}
  \country{}
}
\email{hu.ming.work@gmail.com}
\author{Xiaofei Xie}
\affiliation{%
  \institution{Singapore Management University, Singapore}
  \country{}
}
\email{xfxie@smu.edu.sg}

\author{Yue Cao}
\affiliation{%
  \institution{Nanyang Technological University, Singapore}
  \country{}
}
\email{caoy0033@e.ntu.edu.sg}

\author{Pengyu Zhang}
\affiliation{%
  \institution{East China Normal University, Shanghai, China}
  \country{}
}
\email{zpy71970217@outlook.com}

\author{Yihao Huang}
\affiliation{%
  \institution{National University of Singapore, Singapore}
  \country{}
}
\email{huangyihao22@gmail.com}

\author{Mingsong Chen}
\authornotemark[1]
\affiliation{%
  \institution{East China Normal University, Shanghai, China}
  \country{}
}
\email{mschen@sei.ecnu.edu.cn}

\begin{abstract}

Due to the lack of data auditing techniques for untrusted clients, Federated Learning (FL) is vulnerable to backdoor attacks.
Although various methods have been proposed to protect FL against backdoor attacks, they still
exhibit poor defense performance in extreme data heterogeneity scenarios. 
Worse still, these methods strongly rely on additional datasets, violating the privacy protection requirements of FL.
To overcome the above shortcomings, this paper proposes a novel 
 data-free backdoor defense approach for FL, named  FilterFL,
%
which strives to prevent uploaded client models with backdoor knowledge from participating in the aggregation operation in each FL communication round. 
Based on our knowledge extraction and backdoor filtering schemes using two well-designed Conditional Generative Adversarial Networks (CGANs), FilterFL extracts incremental knowledge learned by a newly updated global model and filters its backdoor components, which can be used to generate one sample that reflects backdoor knowledge for each category. 
If an uploaded local model can confidently classify a generated sample into its target category, the knowledge contributed by the model will be excluded from the aggregation.
In this way, FilterFL can effectively defend against backdoor attacks without using any additional auxiliary data. 
%
%
%
%
%
Comprehensive experiments on well-known datasets demonstrate that, compared with state-of-the-art methods, our approach achieves the best defense performance within various data heterogeneity scenarios.
\end{abstract}


\maketitle
\section{Introduction}

Federated Learning (FL)~\cite{mcmahan2017communication,hu2024fedmut,huang2024federated}, as a promising privacy-aware distributed machine learning paradigm, has been increasingly investigated in various security-critical scenarios, such as healthcare~\cite{cvpr_quande_2021}, autonomous driving~\cite{auto_drive_ijcnn}, and real-time control~\cite{hu2023gitfl}. 
Typically, FL adopts a client-server architecture involving one server and numerous clients, where the server resorts to a global model to enable knowledge sharing among clients without touching their private data.
%
During each FL training round, the server first sends the global model to selected clients for local training and then aggregates all learning from locally trained models to enrich the knowledge of the global model. 
Since the interactions between the server and clients are mainly based on model gradients, FL allows clients to collaboratively train a high-performance model without compromising their data privacy.

Although existing FL methods have the advantage of privacy-preserving knowledge sharing among clients, due to the lack of data auditing techniques for untrusted clients, they are susceptible to various adversarial attacks. 
In particular, FL is vulnerable to backdoor attacks~\cite{bagdasaryan2020backdoor,wang2020attack,xie2020dba}, which aim to inject backdoors into global models by poisoning the local training data or controlling the training processes of local models. 
Specifically, when the global model is poisoned with backdoors, during the inference stage, adversaries can manipulate the model to misclassify input samples with specific patterns (i.e., \emph{triggers}) into their predetermined target categories. 
To deal with backdoor attacks, existing defense methods for traditional Deep Learning (DL) models are mainly based on retraining poisoned models~\cite{liu2018fine, yue2023model} or reversing backdoor triggers~\cite{wang2019neural, li2021neural}, where defenders are required to have additional auxiliary data or access to training datasets~\cite{zhu2020gangsweep, LiMZGAXFZAA24}. Unfortunately, due to the strict data privacy policy imposed by FL, these methods are not suitable for FL scenarios.
%
%
%
%
%
%
%

To defend against backdoor attacks in FL scenarios, various solutions have been investigated, which can be classified mainly into three categories, i.e., similarity-based methods~\cite{blanchard2017machine,xie2018generalized,yin2018byzantine}, Differential Privacy (DP)-based methods~\cite{sun2019can,dwork2014algorithmic}, and clipping-based methods~\cite{nguyen2022flame}. 
Specifically, similarity-based methods can prevent backdoors from being injected into global models by excluding poisoned model parameters with low similarity to others. 
However, in data heterogeneous scenarios, i.e., non-Independent-and-Identically-Distributed (non-IID) scenarios, the model parameters uploaded by different clients vary greatly, reducing or even completely invalidating the defense performance of such similarity-based methods.
As alternatives, DP-based and clipping-based approaches try to mitigate the impacts of backdoors embedded in global models by adding proper noise or clipping parameters, which inevitably degrade the inference accuracy of global models.
Although the above FL defense methods are promising, their implementations are based on various assumptions, e.g., a small proportion of malicious clients~\cite{blanchard2017machine, xie2018generalized} and additional auxiliary data sets~\cite{cao2020fltrust, zhang2022flip}, which are unrealistic in practice. 
Therefore, {\it how to enable unconditional and high-performance defense against backdoor attacks is becoming an urgent issue in FL design.}

Inspired by the work in~\cite{li2021anti,wu2021adversarial}, we define incremental, normal, and backdoor knowledge, respectively, and perform preliminary studies to investigate the characteristic differences between normal and backdoor knowledge.
We found that i) backdoor knowledge is easier to learn than normal knowledge, ii) backdoor knowledge can mislead inference results of a poisoned model toward the specified attack target category, and iii) poisoned models are sensitive to backdoor knowledge toward specific attack target categories.
%
%
Inspired by these observations, we introduce FilterFL, a novel data-free backdoor defense approach for FL based on our proposed knowledge extraction and backdoor filtering schemes.
Specifically,  FilterFL maintains two Conditional Generative Adversarial Networks (CGANs)~\cite{mirza2014conditional} on the server side to filter out backdoor knowledge learned by the global model.

%
In each FL communication round, FilterFL first uses a CGAN to generate one sample for each category, requiring that each generated sample is classified into its designated category by the newly updated global model, but cannot be confidently classified into any category by the old version of the global model. 
This way, the obtained samples can reflect the learned incremental knowledge for each category in this round.
Note that, according to Observation 1, if a malicious client uploads a poisoned model, the generated sample of its target category should contain the backdoor knowledge.
%
%
Then, based on Observation 2 and another CGAN, FilterFL strives to generate one trigger for each category if possible.
By filtering out all the benign classification knowledge with less impact on the classification results of generated samples, the remaining backdoor knowledge for each category imposed by malicious clients can be used to derive the triggers.   
%
%
When dealing with a generated sample as input, according to Observation 3, if an uploaded local model behaves with high classification confidence for the specified category, the model will be considered poisoned and excluded from participating in the update aggregation in this communication round.  
%
%
This paper makes four major contributions as follows: 
\begin{compactenum}[1.]
    \item We make the first attempt to explore the characteristic differences between normal and backdoor knowledge, which can be used to identify backdoor attacks by malicious clients. 
    \item We propose novel knowledge extraction and backdoor filtering schemes, which can extract incremental knowledge learned by the newly updated global model in each FL communication round and obtain its backdoor component. 
    \item We present FilterFL, a novel data-free backdoor defense method for FL,  which can maximize defense performance without resorting to any auxiliary data. 
    \item We conduct extensive experiments to evaluate the effectiveness and adaptability of our approach in varying complex attack scenarios, demonstrating the superiority of our approach against state-of-the-art (SOTA) methods.
\end{compactenum}

\vspace{-0.1in}
\section{Background and Related Work}\label{section:2}

{\bf Federated Learning.} 
Based on a client-server architecture, FL enables knowledge sharing among clients without compromising their data privacy.
Assume that an FL application consists of a server and $N$ clients. Clients iteratively perform collaborative training by imposing their aggregated knowledge on the global model maintained by the server. Typically, each FL training round $t$ involves three phases: i) the server randomly selects $n$ clients and dispatches the global model $G^{t-1}$ to them; ii) each selected client $C_i$ trains $G^{t-1}$ using its local data to obtain a new local model $w_i$ and uploads it to the server; and iii) the server aggregates the models uploaded by all selected clients to obtain a newly updated global model $G^{t}$. 
Similar to the work in ~\cite{bagdasaryan2020backdoor, xie2020dba}, in this paper, we implement our approach on top of FedAvg~\cite{mcmahan2017communication}, which updates the global model by calculating the weighted average of locally trained models.
In other words, our approach updates the global model based on the formula $G^{t}=\sum_{i=1}^{n}w_i/n$.


{\bf Backdoor Attacks on Traditional DL Models.}
As a first attempt to perform backdoor attacks, BadNets~\cite{gu2017badnets} injects backdoors into a DL model by poisoning a subset of its training samples with specific triggers. 
After training using such poisoned samples, the backdoors are formed in the trained model, reflecting the correlation between the triggers and their corresponding target categories specified by adversaries. 
In the inference stage, if a sample is embedded with some trigger, the model will be fooled and misclassified into its specified category.   
Otherwise, the backdoored model will behave as a benign model, which correctly classifies input samples into their true categories. 
%
%
%
Since the quality of triggers determines the attack success ratio, various backdoor attacks have been proposed thereafter, focusing on the improvement of triggers. 
For example, the Blended attack~\cite{chen2017targeted} uses patterns of actual objects rather than simple pixel stripes as triggers. By increasing the transparency of triggers, Blended enhances the invisibility of attacks, making attacks more stealthy.
The Reflect attack~\cite{liu2020reflection} leverages the traces of specular reflection as triggers, making the trigger more invisible. 
In~\cite{turner2019label}, the Clean-Label (CL) attack utilizes adversarial perturbation to blur the normal features of samples of a specific attack target category, forcing models to focus learning attention on triggers during the training phase. 
%
%
%
This way, backdoors can be injected into models without changing data labels, making poisoned data less susceptible to detection by manual review.
%
Rather than poisoning samples with only one individual trigger, various specific-sample attacks~\cite{nguyen2020input, zhang2022neurotoxin} have also been investigated. By generating a specific trigger for each sample, such attacks are more difficult to defend against because of the diversity of triggers. 

%
%

{\bf FL-Oriented Backdoor Attacks.}
According to the injection frequency of backdoor knowledge, 
existing FL-oriented backdoor attack methods~\cite{bagdasaryan2020backdoor, wang2020attack, xie2020dba, zhang2022neurotoxin} ban be mainly classified into two categories, i.e., {\it multiple attacks} and {\it single attacks}. 
For multiple attacks, when a malicious client is selected by the server for local training in one FL training round, it will first tentatively inject target backdoors into its local model and then upload the backdoored model to participate in the knowledge aggregation. Consequently, its target backdoors can be gradually injected into the global model based on the accumulated backdoor knowledge contributed by the uploaded backdoored models. 
However, if an FL application has more benign models than malicious ones, then the backdoor injection process will be very time-consuming and require a large number of FL training rounds. To solve this problem, most multiple attacks rely on controlling a large number of clients to enhance the attacks.
%
%
%
As a different FL attack paradigm, a single attack (a.k.a., a one-shot attack) can inject backdoors into the global model with only one attack by a malicious client. However, to highlight the impact of the malicious client in one aggregation operation, a single attack needs to scale up the parameter weights of its uploaded poisoned model, which greatly decreases the stealthiness of such attacks.
%
To deal with such attacks, defenders can easily identify such a poisoned model with parameter weights significantly larger than the other normal ones and exclude them from the aggregation process~\cite{sun2019can}.
%
Note that since FL continues to update the global model with benign local models after a single attack, the impact of the attack will gradually diminish~\cite {ZhangJCLW23, FangC23}, leading to a low attack success rate. 

{\bf Backdoor Defenses.}
To deal with backdoor attacks on traditional DL models, existing defense methods strive to identify and eliminate backdoors through trigger inversion~\cite{wang2019neural, li2021neural}, model pruning~\cite{liu2018fine}, and retraining~\cite{yue2023model, yyx2024}. 
Although these methods are promising for erasing backdoors from DL models, they cannot be directly applied to FL scenarios. This is mainly because such methods are highly dependent on additional auxiliary training data or access to local training datasets, thus compromising the FL data privacy policy. 
%
To overcome these shortcomings, various FL-specific backdoor defense methods have been investigated, which can be mainly
classified into three categories:
i) similarity-based methods~\cite{blanchard2017machine, xie2018generalized, yin2018byzantine} that eliminate poisoned models with low similarity in parameters to other models~\cite{blanchard2017machine} 
or avoid aggregating abnormal model parameters by changing their aggregation strategies~\cite{xie2018generalized, yin2018byzantine}, 
ii) Differential Privacy (DP)-based methods~\cite{sun2019can,dwork2014algorithmic} that apply a sufficient amount of noise into the global model to eliminate the effects of backdoor attacks, 
and iii) clipping-based methods~\cite{nguyen2022flame} that clip model parameters to limit the influence of clients and global model poisoning.
%
%
Although the above methods take the privacy requirement of FL into account, most of them have their own significant drawbacks. 
Specifically, similarity-based methods suffer from low defense performance in non-IID scenarios due to the divergent learning directions of client models.  
Worse still, when most clients are malicious, the anomaly detection capability of similarity-based methods is significantly weakened, resulting in the neglect of abnormal model parameters.
%
%
%
%
For DP- and clipping-based methods, it is difficult to precisely determine how much noise or how many clippings are required for defenses. 
Consequently, most such methods rely on excessively added noise or clippings, inevitably resulting in deteriorated model accuracy. 

To the best of our knowledge, FilterFL is the first attempt to defend against backdoor attacks for FL utilizing the characteristic differences between normal and backdoor knowledge. 
Without using any additional auxiliary data, FilterFL can maximize the overall defense performance of FL in the data heterogeneity scenario.


%

\vspace{-0.1in}
\section{Our FilterFL approach}\label{section:pre}
\subsection{Threat Model}

{\bf Defense Scope.} The goal of our approach
is to protect horizontal FL (i.e., cross-device FL)~\cite{mcmahan2017communication} against backdoor attacks by identifying poisoned local models uploaded to the server before updating the global model.
In this paper, we assume that attackers cannot control the server or influence benign clients. 
Meanwhile, we do not consider any attacks launched during the communication between the server and clients, such as man-in-the-middle attacks~\cite{callegati2009man}. 
%


{\bf Adversary Capabilities.}
Like most existing FL backdoor attack methods~\cite{xie2020dba,bagdasaryan2020backdoor,nguyen2022flame,zhang2022flip}, an attacker has the following four adversary capabilities:
\ding{182} {\bf multiple clients,}  the attacker can control a subset of participating clients, called malicious clients;
\ding{183} {\bf complete control,} the attacker has complete control over the malicious clients, allowing them to manipulate any specified hyperparameters of FL (e.g., local epochs and learning rates) to train poisoned models;
\ding{184} {\bf attack discretion,} the attacker has the right to decide whether a malicious client conducts an attack or not; 
and 
\ding{185} {\bf collaborative attack,} the attacker can manage multiple malicious clients to perform attacks collaboratively to enhance the overall attack effectiveness.

{\bf Evaluation Metrics.}
We adopt two metrics, i.e., Main task Accuracy (MA) and Attack Success Rate (ASR), to objectively evaluate the effectiveness of a backdoor defense method.
Here, MA denotes the accuracy of inference on clean samples, denoting the benign classification performance of a model. 
In contrast, ASR represents the rate of successful backdoor attacks by poisoned samples, indicating the viciousness of backdoors injected into a model. 
%
Typically, backdoor defenders strive to achieve low ASR and high MA.

\vspace{-0.1in}
\subsection{Preliminary Studies}\label{sec: motivation}
According to the works in ~\cite{li2021anti,souri2022sleeper,zhu2020gangsweep}, when training models, the optimization loss of backdoor tasks decreases faster than that of normal classification tasks.
Inspired by this observation, we reify the knowledge learned by the models based on which we studied the characteristic differences between benign normal knowledge and poisoned backdoor knowledge. 

{\bf Knowledge and Unlearned Knowledge.} 
When training a model on the samples of a specific category, due to the new knowledge learned, its classification capability for this category changes. 
Conceptually, for a model $f_\theta$, its {\it knowledge} about a specific category $l$ can be represented by a sample set, where each sample can be classified into $l$ by $f_\theta$
regardless of its ground truth. 
Inversely, {\it unlearned knowledge} denotes the knowledge not learned by $f_\theta$. It can also be represented by a sample set, where the classification result (i.e., the output soft label) of each sample has similar item weights, indicating that the result of classification by $f_\theta$ is unclear.


%
{\bf Incremental Knowledge.}
We define the {\it incremental knowledge} as the new knowledge learned within one specific
training round by a model. 
Suppose that $f_{\theta_i}$ and $f_{\theta_{i+1}}$ denote two updated versions of model $f_{\theta}$ within the $i^{th}$ and $(i+1)^{th}$ round, respectively. 
The incremental knowledge for Category $l$ learned by $f_{\theta}$ within the $(i+1)^{th}$ round can be represented as the intersection of unlearned knowledge of $f_{\theta_i}$ and learned knowledge of $f_{\theta_{i+1}}$.
%
Due to the privacy issues that client samples are inaccessible to the server in FL, we cannot directly use a sample set to represent the incremental knowledge.
To enable the representation of incremental knowledge within the context of FL,
 we approximate the whole incremental knowledge for Category $l$ learned by $f_{\theta_{i+1}}$ staring from $f_{\theta_i}$ as a representative sample defined as follows:
\begin{equation}\label{equ: pre-1}
    \text{IK}(l;\theta_i;\theta_{i+1}) = \mathop{\arg\min}\limits_{x}\{\text{StD}(f_{\theta_{i}}(x))-f_{\theta_{i+1}}(x)[l]\}.
\end{equation}
Here, $f_{\theta_i}(x)$ is the output of $f_{\theta_i}$ with an input $x$,  $f_{\theta_{i+1}}(x)[l]$ denotes the probability that a sample $x$ belongs to Category $l$ by $f_{\theta_{i+1}}$, and $\text{StD}(\cdot)$ calculates the standard deviation of output soft label weights.
%


{\bf Normal and Backdoor Knowledge.}
Let $D^p$ be a poisoned variant of $D$, where partial samples in $D^p$ contain a trigger targeting Category $t$. 
Assume that $f_{\theta_i}$ is a benign model. 
After training $f_{\theta_i}$ on $D$ and $D^p$ with one more round, respectively, we can get a benign model $f_{\theta^b_{i+1}}$ and a poisoned model $f_{\theta^p_{i+1}}$.
For the benign model $f_{\theta^b_{i+1}}$, we define its {\it normal knowledge} about Category $l$ learned in the $(i+1)^{th}$ round as its incremental knowledge about Category $l$.
For the poisoned model $f_{\theta^p_{i+1}}$, we define its
{\it normal knowledge} and the {\it backdoor knowledge} about Category $l$ learned at the $(i+1)^{th}$ round 
as the intersection and the  difference between 
the incremental knowledge of $f_{\theta^p_{i+1}}$ and $f_{\theta^b_{i+1}}$ for Category $l$, respectively.
Note that $f_{\theta^p_{i+1}}$ has no backdoor knowledge for any category $l$ ($l\neq t$).


\begin{figure}[ht]
   \vspace{-0.15in}
    \centering		
    \includegraphics[width=0.4\textwidth]{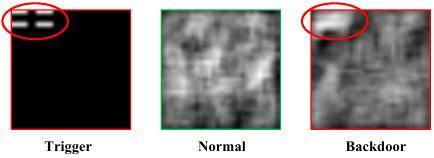}
   \vspace{-0.1in}
    \caption{Comparison between triggers and extracted normal and backdoor knowledge.}
	\label{fig: motivation}
  \vspace{-0.15in}
\end{figure}

{\bf Observation 1.} 
To find the difference between backdoor and normal knowledge, we conducted an experiment on the MNIST dataset $D$, which consists of 10 categories (i.e., 0-9).
In this experiment, we adopted a trigger in the form of the one shown in Figure~\ref{fig: motivation}. 
Assuming that Category 2 is the attack category, to achieve the incremental knowledge derived only by Category 2, we considered a subset of MNIST with samples of Category 2 removed, denoted as $D^{-2}$.
Here, the reason for the removal of the samples of Category 2 is to facilitate the comparison between normal knowledge and backdoor knowledge of Category 2. Note that our approach does not require removing any categories during local model training. 
%
In the experiment, we first trained an initialized model $f_{\theta}$ on $D^{-2}$ with 10 training rounds and obtained a trained model $f_{\theta^{2}}$ that converges. 
Then, we trained the model on $D$ and $D^p$ with one training round, respectively, and obtained a benign model $f_{\theta^{2,b}}$ and a poisoned model $f_{\theta^{2,p}}$.
%
%
In this way, the normal knowledge of $f_{\theta^{2,b}}$ is equivalent to its incremental knowledge, and the backdoor knowledge of $f_{\theta^{2,p}}$ is equal to its incremental knowledge. 
%
According to Equation~\ref{equ: pre-1}, we can use two samples (i.e., $\text{IK}(2;\theta^{2};\theta^{2,b})$ and $\text{IK}(2;\theta^{2};\theta^{2,p})$) as shown in Figure~\ref{fig: motivation} to represent normal and backdoor knowledge for $f_{\theta^{2,b}}$ and $f_{\theta^{2,p}}$, respectively. 
Note that Category 2 samples are used by $D$ and $D^p$ only in the $11^{th}$ round, and both the normal and the backdoor knowledge here are incrementally learned in the $11^{th}$ round.
From this figure, we can find that the extracted backdoor knowledge of $f_{\theta^{2,p}}$ can exhibit the trigger information, while the normal knowledge of $f_{\theta^{2,b}}$ does not reflect any semantic information of Category 2 samples in $D$, indicating that {\it backdoor knowledge is easier to learn than normal knowledge.}





\begin{figure}[ht]
    \vspace{-0.25in}
\centering
	\subfloat[Backdoor knowledge]{
		\centering
		\includegraphics[width=0.22\textwidth]{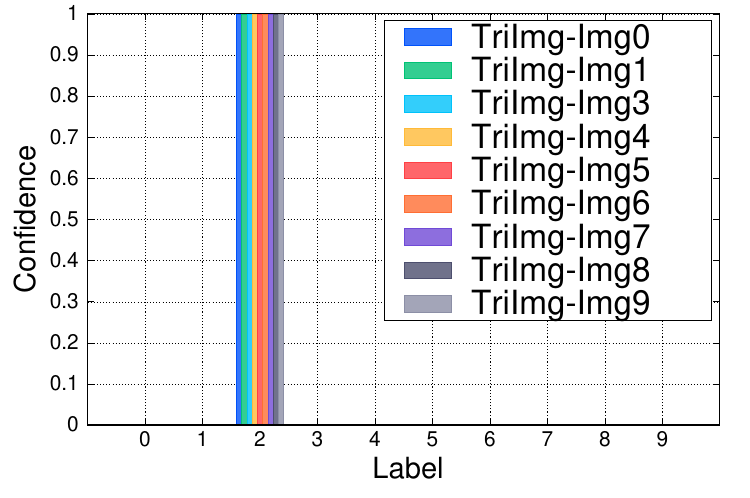}
	}
  \hspace{-0.1in}
    \subfloat[Normal knowledge]{
		\centering
		\includegraphics[width=0.22\textwidth]{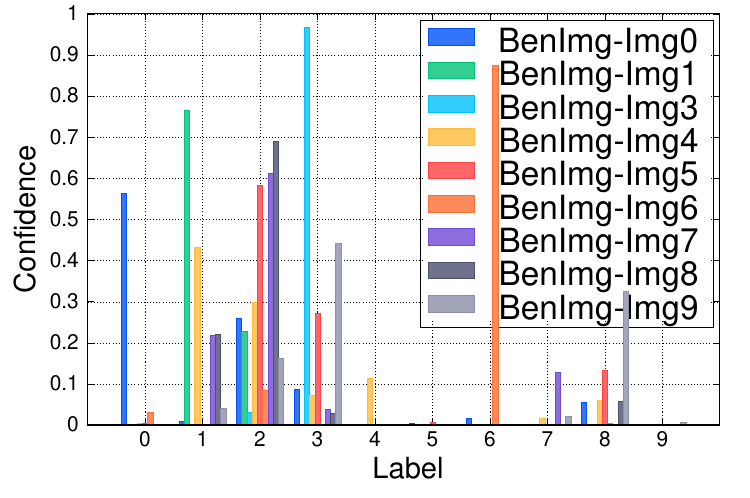}
	}
    \vspace{-0.1in}
 \caption{Impacts of extracted backdoor and normal knowledge on classification results of the poisoned model $f_{\theta^{2,p}}$.
 \vspace{-0.15in}
 } 
 \label{fig: motivation_P_BI}
\end{figure}

\textbf{Observation 2.} 
To understand the impacts of normal and backdoor knowledge on the classification of samples, we conducted an inference experiment on the MNIST dataset.
We superimposed $\text{IK}(2;\theta^2;\theta^{2,p})$ and $\text{IK}(2;\theta^2;\theta^{2,b})$ onto 100 randomly selected samples of MNIST, respectively. 
Figure~\ref{fig: motivation_P_BI} presents the average inference performance of $f_{\theta^{2,p}}$ for the superimposed samples with backdoor and normal knowledge, respectively. 
Here, we use the notions ``TriImg-Img$x$''  and ``BenImg-Img$y$'' to indicate the samples of Category $x$ superimposed by the backdoor and normal knowledge, respectively. 
From this figure, we can find that the inference results of the samples superimposed with backdoor knowledge are uniformly the same as the attack target category. In contrast, the results of their counterparts with normal knowledge are random. 
It means that when superimposed on a sample, {\it backdoor knowledge can effectively mislead inference results towards a specified attack target category, while normal knowledge cannot.}

\begin{figure}[ht]
  \vspace{-0.25in}
\centering
	\subfloat[Poisoned model]{
		\centering
		\includegraphics[width=0.22\textwidth]{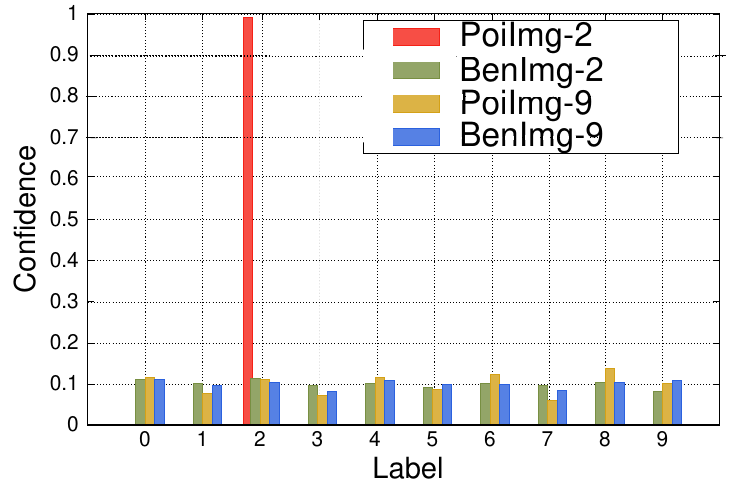}
	}
 \hspace{-0.1in}
    \subfloat[Benign model]{
		\centering
		\includegraphics[width=0.22\textwidth]{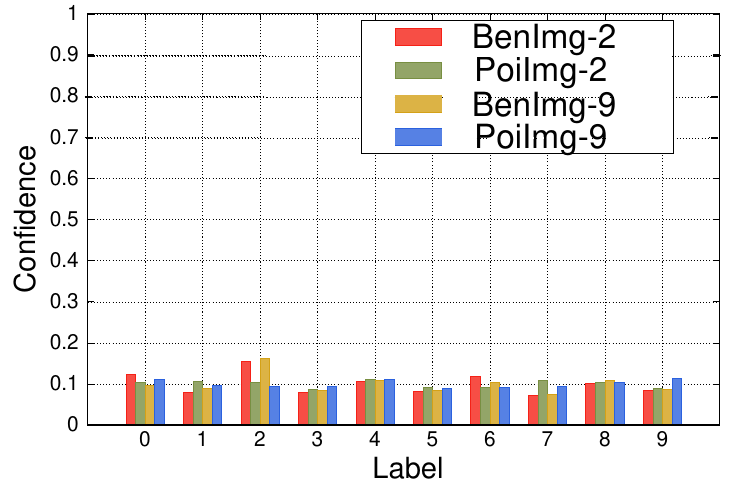}
	}
     \vspace{-0.15in}
 \caption{Classification results of extracted knowledge itself.} 
 \label{fig: motivation_label}
 \vspace{-0.1in}
\end{figure}

\textbf{Observation 3.} 
To investigate the effectiveness of backdoor knowledge in distinguishing poisoned models from benign models, we conducted an experiment on the MNIST dataset using the two models, i.e., $f_{\theta^{2,p}}$ and $f_{\theta^{2,b}}$. 
Figure~\ref{fig: motivation_label} shows the experimental results,
where we use the notions ``PoiImg-$l$'' and ``BenImg-$l$'' to indicate incremental knowledge for Category $l$ extracted from $f_{\theta^{2,p}}$ and $f_{\theta^{2,b}}$, respectively.
Note that, for $f_{\theta^{2,p}}$, the extracted incremental knowledge for Category 2 equals its backdoor knowledge, which can be represented by $\text{IK}(2;\theta^2;\theta^{2,b})$.
From the figure, we can find that $f_{\theta^{2,p}}$ can confidently classify the backdoor knowledge (i.e., ``PoiImg-2'') into the attack target category (i.e., Category 2).
However, for the extracted incremental knowledge, their classification results by both poisoned and benign models are of low confidence.
In other words, {\it compared with benign models, poisoned models are much more sensitive to backdoor knowledge toward specific attack target categories.}


\begin{figure*}[ht]
\centering
\includegraphics[width=0.98\textwidth]{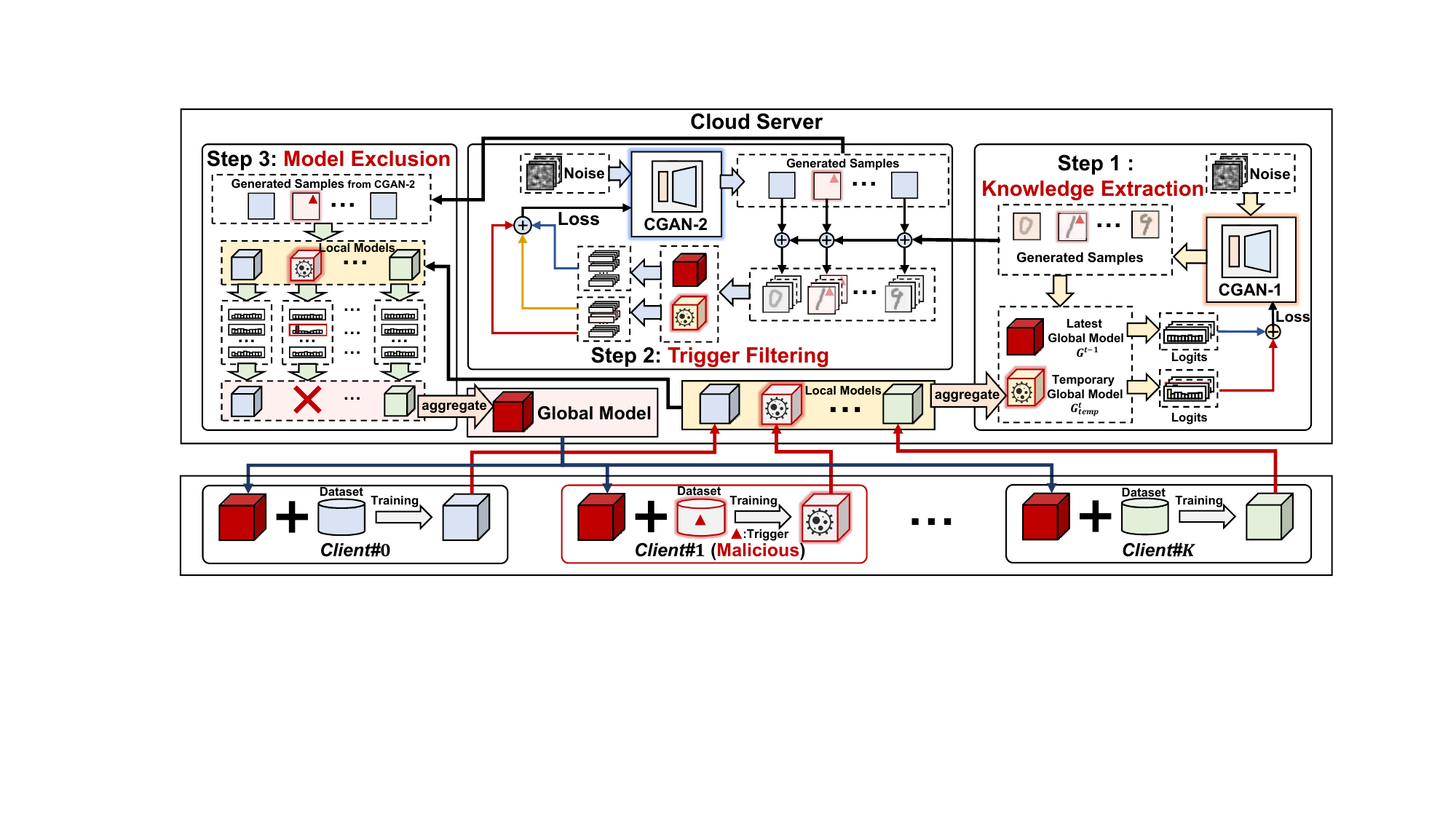}
\vspace{-0.1in}
\caption{Framework and workflow of our FilterFL approach.}
\label{fig: framework}
\vspace{-0.15in}
\end{figure*}

\vspace{-0.1in}
\subsection{Overview of FilterFL}\label{section:4}
Figure~\ref{fig: framework} details the framework and workflow of our FilterFL approach. To facilitate the introduction, we assume that the framework consists of one cloud server and $N$ clients, where $K$ clients are randomly selected in each FL communication round to perform model aggregation.
Typically, backdoor attacks occur during local training of clients, where malicious clients try to inject backdoors into local models by poisoning their local data.
As an example shown in Figure~\ref{fig: framework},  in the $t^{th}$ FL communication round,  $Client\# 2$ is a malicious client that is selected to participate in model aggregation. Applying a red triangle as a trigger on partial samples of local data, a backdoor with Category 1 as its attack target is injected into the local model of $Client\# 2$ during the local training process.  
When $Client\# 2$ is uploaded to the cloud server, it cannot be detected directly by the server, as the cloud server does not have a priori knowledge of the backdoors.
However, according to Observation 3, if we have backdoor knowledge for the attack, FilterFL can effectively identify poisoned models and exclude them from the model aggregation. 
To achieve such backdoor knowledge, FilterFL divides its aggregation process into the following three steps based on two CGANs (i.e., CGAN-1 and CGAN-2).

{\bf Step 1: Knowledge Extraction.}
FilterFL adopts CGAN-1 to figure out the incremental knowledge for each category learned in the round $t$, which is represented by a set of samples. 
To prevent backdoor models from being involved in knowledge aggregation, in Step 1, FilterFL constructs a temporary global model (i.e., $G_{temp}^{t}$) based on all uploaded models. 
%
Based on CGAN-1, FilterFL can effectively extract the incremental knowledge for each category (i.e., a set $\mathbf{S}_{g}^{t}$) between $G_{temp}^{t}$ and the latest global model $G^{t-1}$ obtained in round $t-1$. 
%
%
According to the definition of incremental knowledge, for each sample in $\mathbf{S}_{g}^{t}$, $G_{temp}^{t}$ will correctly classify it into its category, while $G^{t-1}$ cannot classify it into any category with high confidence. 
Note that, according to Observation 1, the obtained incremental knowledge should contain the backdoor knowledge learned by malicious clients (e.g., $Client\#1$).


%


{\bf Step 2: Trigger Filtering.} 
In Step 2, we train CGAN-2 by repetitively superimposing the output samples (i.e.,  $\mathbf{T}_{g}$) of CGAN-2 onto each generated sample from Step 1, respectively,  and feed them into both $G_{temp}^{t}$ and $G^{t-1}$.
When training converges, we can use CGAN-2 to extract backdoor knowledge for each category in the form of samples from incremental knowledge obtained in Step 1. 
At this time, according to Observation 2, if a sample in $\mathbf{T}_{g}$ is superimposed on each generated sample from Step 1,  $G_{temp}^{t}$ will classify all these superimposed samples into the same category instead of their original categories.

%


{\bf Step 3: Model Exclusion.} 
Step 3 conducts inference on the final results of  $\mathbf{T}_{g}$ using each 
uploaded model, respectively. According to Observation 3, if a model can confidently classify a sample in $\mathbf{T}_{g}$ into its corresponding category,  the model cannot participate in the aggregation in round $t$ since it is poisoned. 
By removing all these poisoned models, Step 3 will perform the model aggregation like conventional FL to achieve a new global model $G^{t}$.


To facilitate the understanding of our FilterFL workflow, we consider an FL scenario that performs a 10-classification task, where five clients are randomly selected for model aggregation within each communication round. Assume that in the $t^{th}$ round, $Client\#1$ is malicious with Category 1 as its attack target.
To detect the malicious client and prevent it from participating in model aggregation, in Step 1, FilterFL aggregates the five uploaded models into a temporary global model $G_{temp}^t$.
By extracting incremental knowledge of each category between $G^{t-1}$ and $G^t_{temp}$, we can obtain ten samples to form a sample set $S_g$. 
%
For each sample in $S_g$, in Step 2, FilterFL filters out its normal knowledge and retains only its backdoor knowledge. In this way, we obtain a sample set $\mathbf{T}_g$ that contains ten samples, where $\mathbf{t}_l$ is a newly generated sample for Category $l$ ($l\in[0,9]$). 
Note that since the attack target is Category 1, only the sample $\mathbf{t}_1$ in $\mathbf{T}_g$ contains the backdoor knowledge.
In Step 3, FilterFL applies the five uploaded local models to each sample in $\mathbf{T}_g$ for inference. In this case, the $Client\#1$ model will classify $\mathbf{t}_1$ correctly with high confidence, while the other four local models exhibit low confidence in the classification of all samples in $\mathbf{T}_g$. As a result, FilterFL can identify that the model from $Client\#1$ is poisoned, which should be excluded from the model aggregation for the new global model $G^t$.


\vspace{-0.05in}
\subsection{Knowledge Extraction}\label{section:Knowledge Extraction}
To extract incremental knowledge learned by the temporary global model, we design a CGAN (i.e., CGAN-1) consisting of two discriminator networks (i.e., $\mathbf{d}^1_1$ and $\mathbf{d}^2_1$) and one generator network $\mathbf{g}_1$. 
Assume that we are dealing with the $t^{th}$ communication round during the FilterFL training. At the beginning of the round, we initialize $\mathbf{d}^1_1$ and $\mathbf{d}_1^2$
with the latest global model (i.e., $G^{t-1}$) and the temporary global model (i.e., $G_{temp}^{t}$), respectively, and assign the parameters of $\mathbf{g}_1$ with random weights.  
Note that since $G^0$ is a randomly initialized global model without having any knowledge to compromise the effectiveness of our approach, FilterFL does not require any special defensive measures in the first communication round.
According to Equation~\ref{equ: pre-1}, we train $\mathbf{g}_1$ based on the loss function for each Category $l$ as follows:
\begin{equation}\label{equ: stage 1}
\begin{aligned}
   \mathcal{L}_1(\mathbf{d}_1^1,\mathbf{d}_1^2,x_l)=  \gamma_1 \cdot\text{StD}(\mathbf{d}_1^1(x_l)) 
   + (1-\gamma_1)\cdot (1-\mathbf{d}_1^2(x_l)[l]),  
\end{aligned}
\end{equation}
%
where $x_l$ is a sample generated by $\mathbf{g}_1$ and $\gamma_1$ is a coefficient used to balance the two terms of the equation. 
The training goal of $\mathbf{g}_1$ is to generate a set of samples
indicating the incremental knowledge learned for each category within the $t^{th}$ round, where each sample exhibits different behaviors on the latest model $G^{t-1}$ and the temporary global model $G_{temp}^{t}$.
For each generated sample, $G_{temp}^{t}$ can correctly output the classification result, while $G^{t-1}$ cannot confidently give the classification result. In other words, the standard deviation of the output soft label generated by $G^{t-1}$ is low, indicating that the soft label has similar values for its items. 
At the end of the CGAN-1 training, we can obtain one sample for each category by CGAN-1 to form the incremental knowledge set $\mathbf{S}_g$, whose size is the same as the number of categories in the classification task.
 %
According to Observation 1, backdoor knowledge can be learned more easily than normal knowledge. 
This means that if $G_{temp}^{t}$ is poisoned,  $\mathbf{S}_g$ generated by a well-trained CGAN-1 in the $t^{th}$ round will contain the backdoor knowledge. 
 
%
%
%
%
%

\subsection{Trigger Filtering}\label{section: Trigger Filtering}
When leveraging a sample generated by well-trained CGAN-1, even if it contains backdoor knowledge, it is still hard for the server to figure out poisoned models due to the significant disturbance imposed by accompanying normal knowledge. 
To facilitate the identification of poisoned models from uploaded models, we present a novel trigger filtering mechanism that can effectively filter out backdoor knowledge for each category from the incremental knowledge extracted by CGAN-1, which is represented by a sample. Inspired by Observation 2, we design another CGAN (i.e., CGAN-2) that leverages the difference in impacts on classification results between backdoor and normal knowledge to filter knowledge.
Similar to CGAN-1, CGAN-2 consists of one generator network $\mathbf{g}_2$ and two discriminator networks (i.e., $\mathbf{d}^1_2$ and $\mathbf{d}^2_2$), where the discriminator networks of CGAN-2 have the same initialization and usage as those of CGAN-1.
During the training of CGAN-2, we use $\mathbf{g}_2$ to generate one sample for each Category $l$, denoted as $x_l$.
%
For each $x_l$, 
we superimpose it on each sample in $\mathbf{S}_g$ as follows:
\begin{equation}\label{equ: stage 2-1}
    \mathcal{I}(\mathbf{S}_g,x_l) = \left\{\mathop{\bigcup}\nolimits_{s_i \in \mathbf{S}_g, i\neq l}\{s_i \oplus x_l\}\right\}\cup\{s_l \circleddash x_l\},
\end{equation}
where the operators $\oplus$ and $\circleddash$ represent the addition and subtraction of pixels, respectively.
Specifically, for each Category $l$, we superimpose $x_l$ on each sample $s_i$ where $i\neq l$ and subtract $x_l$ from $s_l$. 
In this way, we obtain a new sample set $\mathbf{I}_l=\mathcal{I}(\mathbf{S}_g,x_l)$, which is of the same size as $S_g$.
Based on the samples in $\mathbf{I}_l$, we design the following loss function to optimize $\mathbf{g}_2$:
%
\begin{equation}\label{equ: stage 2-2}
\begin{aligned}
    \mathcal{L}_2&(\mathbf{d}_2^1, \mathbf{d}_2^2,\mathbf{I}_l,l) = \sum\limits_{I_i \in \mathbf{I}_l}\lambda_1\cdot\frac{\text{StD}(\mathbf{d}_2^1(I_i))}{|\mathbf{I}_l|}\\
    +& \sum\limits_{I_i \in \mathbf{I}_l, i\neq l}\lambda_2\cdot \frac{(1-\mathbf{d}^2_2(I_{i})[l])}{|\mathbf{I}_l|-1}+(1-\lambda_1-\lambda_2)\cdot\mathbf{d}^2_2(I_{l})[l],
\end{aligned}
\end{equation}
where $\lambda_1$ and $\lambda_2$ are balance coefficients.
In Equation~\ref{equ: stage 2-2}, the first term guarantees that the sample $I_i$ belongs to the unlearned knowledge of $G^{t-1}$.
%
The second term aims to maximize the probability of classifying all samples $I_i$ into Category $l$ by $G^{t}_{temp}$, where $I_i\in \mathbf{I}_l$ and $i\neq l$.
When superimposing $x_l$ on all samples $s_i$ with $i\neq l$, the classification results of $s_i$ are altered by $x_l$ to $l$.
Inversely, 
the third term of Equation~\ref{equ: stage 2-2} tries to minimize the probability of classifying the sample $I_l$ into Category $l$. In other words, when subtracting $x_l$ from $s_l$, the classification result of $s_l$ cannot be $l$.
Note that, to optimize $\mathbf{g}_2$, we need to calculate $\mathcal{L}_2$ for all categories. 
After training, CGAN-2 can generate one sample for each category, forming the backdoor knowledge set $\mathbf{T}_g$.
If $l_a$ is an attack target category, the sample $\mathbf{t}_{l_a}$ corresponding to the Category $l_a$ in $\mathbf{T}_g$ can effectively mislead the classification results of other samples in $S_g$.
According to Observation 2, the sample $\mathbf{t}_{l_a}$ contains only the backdoor knowledge.



\subsection{Model Exclusion}
Based on the samples generated in $\mathbf{T}_{g}$, FilterFL can effectively identify poisoned models and remove them from the uploaded models for benign model aggregation.  
Since the cloud server does not have any prior knowledge of attacks, to identify whether a local model is poisoned, FilterFL feeds all samples in $\mathbf{T}_{g}$ to each uploaded model, respectively. 
Assume that a sample $\mathbf{t}_{l_a}$ in $\mathbf{T}_g$ indicates backdoor knowledge corresponding to the attack target category $l_a$. According to Observation 3, if some uploaded model can classify $\mathbf{t}_{l_a}$ into $l_a$ with a confidence degree that exceeds a given threshold $\rho$, we can consider the model as poisoned, as it is particularly sensitive to $\mathbf{t}_{l_a}$. 
Unlike poisoned models, benign models are not sensitive to the sample $\mathbf{t}_{l_a}$, leading to similar item values on their output soft labels. 
Note that if a sample $\mathbf{t}_l$ in $\mathbf{T}_g$ contains no backdoor knowledge, both poisoned and benign models are not sensitive to it. 
Based on the above phenomena, we can identify benign models and use them for aggregation to form a new global model $G^{t}$.

%

\begin{algorithm}[h]
\small
        \caption{Implementation of FilterFL}
        	\label{alg: overview}
\begin{algorithmic}[1]
            \REQUIRE{\rmnum{1}) $C$ $=$ $\{c_0, c_1,...,c_{K}\}$, set of clients; 
            \rmnum{2}) $T$, total number of FL communication rounds; 
            \rmnum{3}) $G^0$, initialized global model;
            \rmnum{4}) $E_1, E_2$, the numbers of training rounds for CGAN-1 and CGAN-2;
            \rmnum{5}) $\mathbf{L}$, category list; 
            \rmnum{6}) $\rho$, filtering threshold;
            }
            \ENSURE{$G^T$, trained global model;}
            \FOR{$t=1$ to $T$}
            \STATE $\widehat{C}$ $\leftarrow$ RandomSelect($C$)
            \STATE The server dispatches $G^{t-1}$ to all clients in $\widehat{C}$
            \FOR{each $c_i \in \widehat{C}$ in parallel}  
            \STATE  $w_i^t$ $\leftarrow$ $c_i$ trains $G^{t-1}$ with local data
            \STATE  $c_i$ uploads $w_i^t$ to the server
            \ENDFOR
            \STATE $W$ $\leftarrow$ ReceiveModel($\widehat{C}$)
            \STATE \textcolor{blue}{/* \textit{Step 1: Knowledge Extraction} */}
            \STATE $G_{temp}^{t}$ $\leftarrow$ $\sum_{w_i^t\in W}w^t_i/|W|$
            \STATE $\{\mathbf{d}_1^1,\mathbf{d}_2^1\}$ $\leftarrow$ $G^{t-1}$, $\{\mathbf{d}_1^2,\mathbf{d}_2^2\}$ $\leftarrow$ $G_{temp}^{t}$, $\{\mathbf{g}_1, \mathbf{g}_2\}$ $\leftarrow$ $\text{random()}$\\
            \FOR{epoch = 1 to $E_1$}
            \STATE $\mathbf{x}$ $\leftarrow$ $\mathbf{g}_1(\mathbf{z}, \mathbf{L})$  \textcolor{blue}{$\ \  //\mathbf{z}$ \textit{is a random vector}}
            \STATE $loss_1$ $\leftarrow 0$
            \FOR{each $x_l \in \mathbf{x}$ }
            \STATE $loss_1$ $\leftarrow$ $loss_1$ $+$ $\mathcal{L}_1$$(\mathbf{d}_1^1, \mathbf{d}_1^2, x_l)$ \textcolor{blue}{\ \  // Equation~\ref{equ: stage 1}}
            \ENDFOR
            \STATE $\mathbf{g}_1$$\leftarrow$ optimize($\mathbf{g}_1$, $loss_1$)
            \ENDFOR
            \STATE $\mathbf{S}_{g}$ $\leftarrow$ $\mathbf{g}_1(\mathbf{z}, \mathbf{L})$
            \STATE \textcolor{blue}{/* \textit{Step 2: Trigger Filtering} */}
            \FOR{epoch = 1 to $E_2$}
            \STATE $\mathbf{x}$ $\leftarrow$ $\mathbf{g}_2(\mathbf{z}, \mathbf{L})$ \textcolor{blue}{$\ \  //\mathbf{z}$ \textit{is a random vector}}
            \STATE $loss_2$ $\leftarrow 0$
            \FOR {each $x_l\in \mathbf{x}$}
            \STATE$\mathbf{I}_l $ $\leftarrow$ $\mathcal{I}(\mathbf{S}_g, x_l)$ \textcolor{blue}{\ \  // Equation~\ref{equ: stage 2-1}}             
            \STATE $loss_2$ $\leftarrow$ $loss_2$ $+$ $\mathcal{L}_2$$(\mathbf{d}_2^1, \mathbf{d}_2^2,\mathbf{I}_l,l)$ \textcolor{blue}{\ \  // Equation~\ref{equ: stage 2-2}}
            \ENDFOR
            \STATE $\mathbf{g}_2$$\leftarrow$ optimize($\mathbf{g}_2$, $loss_2$)
            \ENDFOR
            \STATE  $\mathbf{T}_g $ $\leftarrow$ $\mathbf{g}_2(\mathbf{z}, \mathbf{L})$  
            \STATE \textcolor{blue}{/* \textit{ Step 3: Model Exclusion} */}
            \STATE $W_p$ $\leftarrow \{\}$
            \FOR{each $w_i \in W$}
            \FOR{each $\mathbf{t}_l \in \mathbf{T}_g$}
            \STATE $\mathbf{y}$ $\leftarrow$ $w_i(\mathbf{t}_l)$
            \IF{$\mathop{\arg\max}_{index}$ $(\mathbf{y}) = l$ and $\mathbf{y}[l] > \rho$}
            \STATE $W_p \leftarrow $ $W_p \cup \{w_i\}$         
            \ENDIF
            \ENDFOR
            \ENDFOR
            \STATE $W^{\prime} \leftarrow W - W_p$
            \STATE  $G^{t}$  $\leftarrow$ $\sum_{w_i^t \in W^{\prime}}w_i^t/|W^{\prime}|$
            \ENDFOR
            \end{algorithmic}
\end{algorithm}

\subsection{Implementation of FilterFL}

Algorithm~\ref{alg: overview} details the implementation of FilterFL. 
Lines 2-7 describe the local training process of clients within the $t^{th}$ FL communication round. 
Specifically, in Lines 2-3, the server randomly selects a subset of clients from all clients and dispatches the latest global model $G^{t-1}$ to them.
In Lines 4-6, each selected client trains its model with its local data and uploads the trained local model to the server.
In Line 7, the server receives a set of local models $W$ for the following model aggregation. 
During local training, malicious clients can manage the training processes to achieve poisoned local models, which will result in a poisoned global model.  To defend against such backdoor attacks, FilterFL divides its knowledge aggregation process into three steps, as shown in Lines 9-35. 
Lines 8-17 depict the knowledge extraction step, which figures out the incremental knowledge learned in the $t^{th}$ communication round. 
In Line 9, FilterFL first constructs a temporary global model $G_{temp}^{t}$ by aggregating all uploaded models. 
In Line 10, we initialize the discriminator networks of CGAN-1 and CGAN-2 using $G^{t-1}$ and $G_{temp}^{t}$, respectively, and randomly initialize the CGAN generator networks. 
%
%
Lines 11-16 present the training process of $\mathbf{g}_1$.
In Line 12, we use $\mathbf{g}_1$ to generate one sample for each category and use $\mathbf{x}$ to save all these generated samples. 
Lines 13–15 calculate $loss_1$ based on the loss function defined in Equation~\ref{equ: stage 1}, and use it to optimize the training of $\mathbf{g}_1$ in Line 16. 
%
%
At the end of Step 1, Line 17 obtains a set of samples denoted as $\mathbf{S}_{g}$, where each sample indicates incremental knowledge for a specific category. 
Lines 18-26 detail the implementation of our trigger filtering step.
Similar to the training in Lines 12-16, Lines 19-25 show the training process of $\mathbf{g}_2$, where Line 23 forms a new set of samples (i.e., $\mathbf{I}_l$) based on Equation~\ref{equ: stage 2-1}. 
Based on the loss value calculated in Line 24 following Equation~\ref{equ: stage 2-2}, Line 25 optimizes the training of $\mathbf{g}_2$. At the end of Step 2, Line 26 obtains a set of samples denoted as $\mathbf{T}_{g}$, where each sample may contain backdoor knowledge for a specific category if such information is injected into any uploaded model. 
%
Lines 27-35 detail the implementation of our model exclusion step.
In Line 28, we initialize an empty set $W_p$ to collect poisoned models. 
For each model $w_i$ in $W$, in Line 31, FilterFL feeds each sample in $\mathbf{T}_{g}$ into $w_i$ and records the classification result as $\mathbf{y}$.
In Line 32, FilterFL determines whether $w_i$ is poisoned based on the item values of $\mathbf{y}$  according to our Observation 3. 
If yes, Line 33 adds the poisoned models to $W_p$.
Line 34 obtains a set of benign models $W^{\prime}$ by excluding all poisoned models collected in $W$. 
At the end of Step 3, Line 35 aggregates all benign models in $W^{\prime}$ to form a new global model $G^{t}$.

\vspace{-0.05in}
\section{Experiments}\label{section:5}
To evaluate the effectiveness of FilterFL, we implemented it on top of Pytorch (version 1.13.0). All the experiments were conducted on a workstation equipped with an Ubuntu operating system, one Intel i7-13700K CPU, 64GB memory, and one NVIDIA GeForce RTX4090 GPU. 
For all the experiments, we assumed that there are a total of 100 clients involved in FL, and in each round, only 10\% of the clients are randomly selected for local training and aggregation.

\subsection{Experimental Setup}\label{sec: setup}

\begin{table}[ht]
\vspace{-0.1in}
\centering
\caption{Experimental settings of datasets and classifiers.}
\vspace{-0.1in}
\label{table: detailed dataset}
\addtolength{\tabcolsep}{-2pt}
\scalebox{0.90}{
\begin{tabular}{c|c|c|c|c}
\toprule
\multirow{2}{*}{Dataset} & \multirow{2}{*}{Label} & \multirow{2}{*}{Image Size} & Dataset Size & Default \\
 &  &  & \multicolumn{1}{l|}{(Training / Testing)} & Classifier \\ \midrule
 \rowcolor{gray!15}
 MNIST & 10 & 28$\times$28$\times$1 & 60000 / 10000 & CNN \\
CIFAR-10 & 10 & 32$\times$32$\times$3 & 50000 / 10000 & ResNet-8 \\
 \rowcolor{gray!15}GTSRB & 43 & 32$\times$32$\times$3 & 39252 / 12630 & ResNet-8 \\
CIFAR-100 & 100 & 32$\times$32$\times$3 & 50000 / 10000 & ResNet-18 \\
 \rowcolor{gray!15}Tiny-ImageNet & 200 & 64$\times$64$\times$3 & 100000 / 10000 & ResNet-18 \\ \bottomrule
\end{tabular}
}
\vspace{-0.1in}
\end{table}

{\bf Dataset and Model Settings.}
We investigated five classical datasets (i.e.,  MNIST~\cite{lecun1995learning}, CIFAR-10, CIFAR-100, GTSRB~\cite{stallkamp2012man}, and Tiny-ImageNet~\cite{russakovsky2015imagenet}) in the experiments. 
In this paper, each FL training result on MNIST involves 100 communication rounds, while each FL training on the other four datasets requires 2000 communication rounds. 
%
Table~\ref{table: detailed dataset} shows the settings of these datasets and their corresponding
classifiers. 
For MNIST, we adopted a self-defined model that consists of two convolutional layers and two fully connected layers. For the CIFAR-10 and GTSRB datasets, we used ResNet-8~\cite{he2016identity} as classifiers, and for CIFAR-100 and Tiny-ImageNet, we used ResNet-18 as classifiers, where ResNet-8 is a lightweight version of ResNet-18. Furthermore, we used VGG-16~\cite{simonyan2014very} and ResNet-34 to evaluate the performance of our approach in Section~\ref{sec: adaptability}. 
Note that the CGANs used by FilterFL are built based on the model structure proposed in~\cite{mirza2014conditional}. 


{\bf Client Data  Settings.}
In the experiments, we considered various IID and non-IID scenarios to evaluate the adaptivity and performance of our approach with different data distributions. 
Specifically, we adopted the Dirichlet distribution~\cite{hsu2019measuring} to control the heterogeneity among the local data of clients, where a smaller coefficient $\alpha$ indicates a higher data heterogeneity.  By default, we set $\alpha$ to 0.5 for the following experiments. 


\begin{table*}[ht]
\centering
\caption{Defense performance comparison against backdoor attacks 
within a non-IID scenario ($\alpha=0.5$).
}
\vspace{-0.1in}
\label{table: datasets}
\addtolength{\tabcolsep}{2.5pt}
\scalebox{0.90}{
\begin{tabular}{c|cc|cc|cc|cc|cc}
\toprule
\multicolumn{1}{c|}{\multirow{2.5}{*}{Method}} & \multicolumn{2}{c|}{MNIST} & \multicolumn{2}{c|}{CIFAR-10} & \multicolumn{2}{c|}{CIFAR-100} & \multicolumn{2}{c|}{GTSRB} & \multicolumn{2}{c}{Tiny-ImageNet} \\ \cmidrule(l){2-11} 
\multicolumn{1}{c|}{} & MA(\%) & ASR(\%) & MA(\%) & ASR(\%) & MA(\%) & ASR(\%) & MA(\%) & ASR(\%) & MA(\%) & ASR(\%) \\ \midrule
No Attack & 98.38 & 0.22 & 85.52 & 2.51 & 54.31 & 0.50 & 95.63 & 0.20 & 36.45 & 0.41 \\
No Defense & 98.46 & 99.83 & 84.92 & 92.80 & 56.36 & 99.52 & 95.51 & 99.60 & 36.48 & 99.89 \\ \midrule
\rowcolor{gray!15}Krum~\cite{blanchard2017machine} & 93.05 & 1.23 & 70.66 & 7.22 & 46.27 & 0.64 & 94.30 & 0.22 & 24.12 & 0.57 \\
COMED~\cite{yin2018byzantine} & 98.14 & 0.62 & 84.08 & 61.30 & 51.81 & 99.43 & 95.46 & 91.59 & 30.87 & 99.68 \\
\rowcolor{gray!15}RLR~\cite{ozdayi2021defending} & 93.85 & 1.84 & 76.93 & 8.30 & 26.64 & 42.25 & 94.13 & 34.26 & 20.73 & 68.87 \\
Trimmed-Mean~\cite{xie2018generalized} & 98.19 & 0.69 & 84.59 & 64.54 & 52.04 & 99.47 & \textbf{95.70} & 91.59 & 31.07 & 99.87 \\
\rowcolor{gray!15}DP~\cite{sun2019can} & 90.61 & 85.66 & 71.23 & 84.21 & 41.56 & 89.70 & 90.44 & 88.80 & 26.12 & 81.19 \\
FLAME~\cite{nguyen2022flame} & 98.16 & 0.37 & 81.85 & 93.78 & 51.52 & 98.24 & 94.76 & 66.78 & 34.96 & 89.15 \\
\rowcolor{gray!15}FLTrust~\cite{cao2020fltrust} & 98.35 & 0.31 & 82.55 & 20.53 & 53.45 & 29.69 & 95.45 & 25.03 & 35.28 & 33.67 \\
FLIP~\cite{zhang2022flip} & 98.43 & 1.32 & 85.29 & 10.23 & 54.32 & 22.32 & 95.36 & 12.56 & 35.12 & 28.71 \\
\rowcolor{gray!15}FLDetector~\cite{zhang2022fldetector} & 97.51 & 0.40 & 82.25 & 10.17 & 51.98 & 7.61 & 94.49 & 6.63 & 30.27 & 86.71 \\
FedREDefense~\cite{yueqi2024fedredefense} & 98.52 & 0.38 & 81.32 & 2.50 & 51.46 & \textbf{0.61} & 94.28 & 10.21 & 34.15 & 12.03 \\
\rowcolor{gray!45}FilterFL (Ours) & \textbf{98.55} & \textbf{0.31} & \textbf{85.46} & \textbf{2.48} & \textbf{54.95} & 0.65 & 95.45 & \textbf{0.14} & \textbf{35.32} & \textbf{0.57} \\ \bottomrule
\end{tabular}
}
\vspace{-0.15in}
\end{table*}

{\bf Attack Settings.}
In the experiments, we mainly investigated BadNets~\cite{gu2017badnets} as the default attack method for FL, which has been widely investigated to perform backdoor attacks within FL scenarios. 
When dealing with the FL version of BadNets, we followed the same attack settings used in~\cite{bagdasaryan2020backdoor,xie2020dba} and applied a white stripe in the upper left corner of the samples as the default trigger pattern.
To evaluate the performance of FilterFL against SOTA FL backdoor attacks, we also investigated Blended~\cite{chen2017targeted}, Reflect~\cite{liu2020reflection}, 
Clean-Label~\cite{turner2019label}, 
DBA~\cite{xie2020dba}, 
Neurotoxin~\cite{zhang2022neurotoxin}, 
A3FL~\cite{ZhangJCLW23}
and F3BA~\cite{FangC23} to launch backdoor attacks following their original setups, respectively. 
We perform the multiple attack type (denoted by ``M'') and the single attack type (denoted by ``S'') to deploy the above attacks in FL scenarios, respectively.
By default, we used the multiple attack type to deploy attacks,
which means that all malicious clients continuously strive to inject backdoors into the global model. 
Here, we used a variable $\eta$ ($\eta=0.1$ by default) to denote the proportion of malicious clients among all clients. 
For each malicious client, we assumed that a subset of its local data was poisoned, where the poisoning rate was 0.3 by default. 

{\bf FL Defense Baseline Settings.}
We compared FilterFL with ten SOTA baseline methods, i.e., Krum~\cite{blanchard2017machine}, COMED~\cite{yin2018byzantine}, RLR~\cite{ozdayi2021defending}, Trimmed-Mean~\cite{xie2018generalized}, DP~\cite{sun2019can}, FLAME~\cite{nguyen2022flame}, FLTrust~\cite{cao2020fltrust}, FLIP~\cite{zhang2022flip}, FLDetector~\cite{zhang2022fldetector} and FedREDefense~\cite{yueqi2024fedredefense}. 
Specifically, {\bf Krum} uses the geometric distance of parameters between models to choose the model that is most similar to the others as the new global model. 
{\bf COMED} changes the aggregation rules to update the global model with the median instead of the average. 
In {\bf RLR}, which reverses the update directions of parameters that differ from those of the majority, we set the maximum number of reversals $\theta$ to 4.
%
In {\bf Trimmed-Mean} that removes the extreme values of parameters when calculating the average for aggregation,  we set the number of extreme values $k$ to 3. 
In {\bf DP} that adds random noise to models to reduce the influence of attacks,  we set the variance $\sigma$ of Gaussian noise to 0.015. 
In {\bf FLAME} that uses HDBSCAN to cluster outliers and confirm the degree of noise and clippings, we calculated the cosine similarity distance of HDBSCAN, and set $min\_cluster\_size$ and $min\_samples$ to 6 and 1, respectively. 
%
In {\bf FLTrust}, which detects poisoned models by training a standard model on the server as a reference, we randomly selected 100 training samples from the original dataset to compose an additional dataset on the server. 
By setting aside half of the training batch as augmented samples for each benign client, {\bf FLIP} inverts the triggers using Neural Cleanse~\cite{wang2019neural} and uses them to train benign clients to reduce the impact of backdoors.
Following the original settings of  {\bf FLDetector} that leverages global models to estimate local gradients, we detected malicious clients by analyzing training trajectories in 10 rounds after 50 cold start rounds.
%
For {\bf FedREDefense}, we used its original reconstruction hyperparameters to identify malicious clients based on discrepancies in model update reconstruction errors.



\vspace{-0.1in}
\subsection{Defense Performance Comparison}
We compared FilterFL with the ten baselines on five well-known datasets (i.e., MNIST, CIFAR-10, CIFAR-100, GTSRB, and Tiny-ImageNet, respectively) within one IID and three non-IID scenarios (i.e., $\alpha=0.1$, $0.5$, and $1.0$, respectively). 
Due to space limitations, please refer to Table~\ref{table: complete} in the Supplementary Material for complete performance comparisons. 

{\bf Performance on Different Datasets.}
Table~\ref{table: datasets} compares
the Main task Accuracy (MA) and Attack Success Rate (ASR) between our approach and the 
 ten baselines on MNIST, CIFAR-10, CIFAR-100, GTSRB, and Tiny-ImageNet datasets within a non-IID scenario ($\alpha=0.5$), respectively.
From this table, we can observe that our approach can achieve the lowest ASR and the best MA on the MNIST, CIFAR-10, and Tiny-ImageNet datasets, respectively.
For CIFAR-100, FilterFL can achieve the best MA and the third-lowest ASR.
In this case, although Krum and FedREDefense outperform FilterFL by $0.01\%$ and $0.04\%$ in terms of ASR, respectively, 
FilterFL can achieve much better MA (i.e., by $8.68\%$ and $2.97\%$, respectively) than Krum and FedREDefense.
%
For GTSRB, FilterFL stood out with the lowest ASR among all baselines. 
Note that in this case, although the MA of FilterFL is slightly lower ($0.25\%$) than the best MA obtained by Trimmed-Mean, FilterFL significantly outperforms Trimmed-Mean by $91.45\%$ in terms of ASR.
Moreover, we can observe that most baselines perform well on MNIST but exhibit
unsatisfactory performance on the other datasets, where the samples are images with three channels. 
In summary, compared with the SOTA defenses, our approach can minimize ASR while achieving competitive MA for different datasets.




\begin{table}[h]
\vspace{-0.1in}
\centering
\caption{Defense performance comparison against backdoor attacks on CIFAR-10 considering different data distributions.}
\vspace{-0.1in}
\label{table: Distributions}
\addtolength{\tabcolsep}{-3.5pt}
\scalebox{0.8}{
\begin{tabular}{c|cc|cc|cc|cc}
\toprule
\multicolumn{1}{c|}{\multirow{2.5}{*}{Method}} & \multicolumn{2}{c|}{IID} & \multicolumn{2}{c|}{$\alpha=1.0$} & \multicolumn{2}{c|}{$\alpha=0.5$} & \multicolumn{2}{c}{$\alpha=0.1$} \\ \cmidrule(l){2-9} 
\multicolumn{1}{c|}{} & \multicolumn{1}{c}{MA(\%)} & \multicolumn{1}{c|}{ASR(\%)} & \multicolumn{1}{c}{MA\%} & \multicolumn{1}{c|}{ASR(\%)} & \multicolumn{1}{c}{MA(\%)} & \multicolumn{1}{c|}{ASR(\%)} & \multicolumn{1}{c}{MA(\%)} & \multicolumn{1}{c}{ASR(\%)} \\ \midrule
No Attack & 86.53 & 2.31 & 86.59 & 2.06 & 85.52 & 2.51 & 74.17 & 6.50 \\
No Defense & 84.33 & 91.73 & 85.22 & 88.92 & 84.92 & 92.80 & 77.12 & 97.04 \\\midrule
\rowcolor{gray!15}Krum & 84.78 & 2.76 & 77.83 & 3.26 & 70.66 & 7.22 & 39.08 & \textbf{3.74} \\
COMED & 86.28 & 10.76 & 85.95 & 30.22 & 84.08 & 61.30 & 64.37 & 98.59 \\
\rowcolor{gray!15}RLR & 78.84 & 3.89 & 80.58 & \textbf{0.80} & 76.93 & 8.30 & 68.66 & 89.36 \\
T-Mean & 86.20 & 14.32 & 86.10 & 30.50 & 84.59 & 64.54 & 66.81 & 97.08 \\
\rowcolor{gray!15}DP & 71.13 & 84.48 & 71.95 & 84.05 & 71.23 & 84.21 & 56.73 & 89.35\\
FLAME & 85.56 & 10.85 & 84.24 & 87.17 & 81.85 & 93.78 & 66.89 & 98.86 \\
\rowcolor{gray!15}FLTrust &84.42 & 2.88 & 84.56 & 3.83 & 82.55 & 20.53& 69.63 & 65.72 \\
FLIP & 86.65 & 2.69 & 86.11 & 4.12 & 85.29 & 10.23 & 72.71 & 24.52\\
\rowcolor{gray!15}FLDetector & 86.23 & 2.71 & 84.30 & 3.62 & 82.25 & 10.17 & 69.86 & 80.57\\
FedREDefense & 85.14 & 2.35 & 85.52 & 2.37 & 81.32 & 2.50 & 65.56 & \textbf{7.42} \\
\rowcolor{gray!45}FilterFL (Ours) & \textbf{86.66} & \textbf{2.33} & \textbf{86.70} & 2.27 & \textbf{85.46} & \textbf{2.48} & \textbf{72.63} & 7.42 \\ \bottomrule
\end{tabular}
}
\vspace{-0.05in}
\end{table}

{\bf Impact of  Data Heterogeneity.}
Table~\ref{table: Distributions} investigates the impact of data heterogeneity on defense performance, comparing MA and ASR between FilterFL and the ten baselines on the CIFAR-10 dataset, considering various data distributions. 
From this table, we can find that FilterFL outperforms other baselines in most cases. Especially, FilterFL can approximate the best performance shown in the ``No Attack'' case for all distributions. 
However, most baselines only perform well in the IID scenario. For example, Krum can achieve an MA of $84.78\%$ in the IID case, while its MA is $39.08\%$ in the non-IID case with $\alpha=0.1$. 
This is because Krum updates the global model by selecting only one client instead of aggregating multiple models. Therefore, it can accurately exclude poisoned models from the FL aggregation process, resulting in a satisfactory ASR in various scenarios. 
However, we find that the MA of Krum decreases significantly along with the decrease of the value of $\alpha$. Note that when $\alpha=0.1$, it is difficult for Krum to converge, resulting in a significantly low MA. 
Although FedREDefense and our FilterFL can simultaneously achieve the lowest ASR when $\alpha=0.1$, the MA of FedREDefense is $7.07\%$ lower than ours.

%


\begin{table}[t]
\centering
\caption{Defense performance comparison against BadNets (M) attacks with different proportions of malicious clients.}
\vspace{-0.1in}
\label{table: times}
\addtolength{\tabcolsep}{-3.5pt}
\scalebox{0.8}{
\begin{tabular}{c|cc|cc|cc|cc}
\toprule
\multicolumn{1}{c|}{\multirow{2.5}{*}{Method}} & \multicolumn{2}{c|}{$\eta=0.1$} & \multicolumn{2}{c|}{$\eta=0.2$} & \multicolumn{2}{c|}{$\eta=0.5$} & \multicolumn{2}{c}{$\eta=0.8$} \\ \cmidrule(l){2-9} 
\multicolumn{1}{c|}{} & \multicolumn{1}{c}{MA(\%)} & \multicolumn{1}{c|}{ASR(\%)} & \multicolumn{1}{c}{MA(\%)} & \multicolumn{1}{c|}{ASR(\%)} & \multicolumn{1}{c}{MA(\%)} & \multicolumn{1}{c|}{ASR(\%)} & \multicolumn{1}{c}{MA(\%)} & \multicolumn{1}{c}{ASR(\%)} \\ \midrule 
No Attack & 98.38 & 0.22 & 98.38 & 0.22 & 98.38 & 0.22 & 98.38 & 0.22 \\
No Defense & 98.46 & 99.83 & 98.64 & 99.98 & 98.63 & 100.00 & 98.60 & 100.00 \\ \midrule
\rowcolor{gray!15}Krum & 93.05 & 1.23 & 94.11 & 0.75 & 88.02 & 95.38 & 97.09 & 99.84 \\
COMED & 98.14 & 0.62 & \textbf{98.32} & 97.51 & 98.58 & 99.99 & 98.51 & 100.00 \\
\rowcolor{gray!15}RLR & 93.85 & 1.84 & 98.29 & 99.33 & 97.80 & 99.31 & 98.10 & 100.00 \\
T-Mean & 98.19 & 0.69 & 98.31 & 99.55 & \textbf{98.61} & 100.00 & \textbf{98.59} & 100.00 \\
\rowcolor{gray!15}DP  & 90.61 & 85.66 & 90.23 & 84.21 & 86.56 & 95.70 & 84.44 & 99.80\\
FLAME & 98.16 & 0.37 & 98.16 & 2.33 & 98.45 & 100.00 & 98.44 & 100.00 \\
\rowcolor{gray!15}FLTrust & 98.35 & 0.31 & 98.27 & 1.43 & 94.23 & 2.25& 92.12 & 2.72 \\
FLIP & 98.43 & 1.32 & 98.21 & 4.56 & 97.19 & 20.68 & 98.87 & 63.98\\
\rowcolor{gray!15}FLDetector & 97.51 & 0.40 & 97.19 & 0.57 & 94.29 & 3.62 & 95.81 & 76.55\\
FedREDefense & 98.52 & 0.38 & 98.20 & 0.25 & 94.98 & 0.37 & 82.13 & 0.91\\
\rowcolor{gray!45}FilterFL (Ours) & \textbf{98.55} & \textbf{0.31} & \textbf{98.32} & \textbf{0.21} & 94.96 & \textbf{0.33} & 93.82 & \textbf{0.85} \\ \bottomrule
\end{tabular}
}
\vspace{-0.15in}
\end{table}

{\bf Impact of the Proportion of Malicious Clients.}
To explore the performance of our proposed FilterFL against backdoor attacks involving different proportions of malicious clients,
we conducted experiments with different $\eta$, i.e., $\eta=0.1$, $0.2$, $0.5$, and $0.8$, respectively.
Note that, as shown in Table~\ref{table: datasets}, most baselines only work well on the MNIST dataset. 
To fairly understand the impact of the proportion of malicious clients, the experiments here were conducted only on MNIST. 
Table~\ref{table: times} presents the experimental results on MNIST within a non-IID scenario ($\alpha=0.5$).
From this table, we can observe that as the proportion of malicious clients increases, the effectiveness of most baselines against backdoor attacks diminishes, while FilterFL can always achieve the best backdoor defense performance with the lowest ASR. 
For example, when $\eta=0.2$, Krum, FLAME, FLTrust, FLIP, FLDetector, FedREDefense, and our FilterFL can achieve an ASR lower than $2.5\%$. Moreover, when $\eta=0.5$ or $0.8$, only FLTrust, FedREDefense, and our approach can resist backdoor attacks. 
Notably, our approach achieves much better ASR than FLTrust when $\eta\geq 0.2$.
Meanwhile, our FilterFL achieves a better MA than FedREDefense when $\eta=0.8$.
The reason why FLTrust achieves superior defense performance is that FLTrust detects poisoned models by using an independent standard model on the server side, whose performance is not determined by the number of malicious clients. 
Similarly, FedREDefense detects malicious clients based on the characteristics of the model update, which is not disturbed by the high proportion of malicious clients.
We can also find that the defense performance of FilterFL gets slightly worse as $\eta$ increases. This is because the more malicious clients involved in FL training, the more poisoned models will be excluded from knowledge aggregation within one training round, thus inevitably affecting the generalization ability of the global model.

\begin{table}[ht]
\vspace{-0.05in}
\centering
\caption{Performance of FilterFL on FL scenarios with different sampling rates on CIFAR-10.}
\vspace{-0.1in}
\label{table: sampling rate}
\addtolength{\tabcolsep}{-1.5pt}
\scalebox{0.8}{
\begin{tabular}{@{}c|cc|cc|cc|cc@{}}
\toprule
\multirow{2.5}{*}{\begin{tabular}[c]{@{}c@{}}Sampling\\ Rate\end{tabular}} & \multicolumn{2}{c|}{IID} & \multicolumn{2}{c|}{$\alpha=1.0$} & \multicolumn{2}{c|}{$\alpha=0.5$} & \multicolumn{2}{c}{$\alpha=0.1$} \\ \cmidrule(l){2-9} 
 & MA(\%) & ASR(\%) & MA\% & ASR(\%) & MA(\%) & ASR(\%) & MA(\%) & ASR(\%) \\ \midrule
\rowcolor{gray!15}5\% & 86.43 & 2.21 & 85.24 & 2.30 & 85.02 & 2.37 & 68.49 & 7.52 \\
10\% & 86.66 & 2.33 & 86.70 & 2.27 & 85.46 & 2.48 & 72.63 & 7.42 \\
\rowcolor{gray!15}50\% & 86.97 & 2.35 & 87.02 & 2.28 & 85.51 & 2.41 & 73.16 & 7.21 \\
100\% & 88.41 & 2.16 & 87.50 & 2.33 & 85.69 & 2.38 & 74.33 & 6.97 \\ \bottomrule
\end{tabular}
}
\vspace{-0.2in}
\end{table}

\begin{table*}[h]
\centering
\caption{Performance of FilterFL against SOTA FL backdoor attacks within a non-IID scenario ($\alpha=0.5$).}
\vspace{-0.1in}
\addtolength{\tabcolsep}{-4pt}
\scalebox{0.76}{
\label{table: attack}
\begin{tabular}{c|cccc|cccc|cccc|cccc|cccc}
\toprule
 & \multicolumn{4}{c|}{MNIST} & \multicolumn{4}{c|}{CIFAR-10} & \multicolumn{4}{c|}{CIFAR-100} & \multicolumn{4}{c|}{GTSRB} & \multicolumn{4}{c}{Tiny-ImageNet} \\ \cmidrule(l){2-21} 
Attack & \multicolumn{2}{c|}{No Defense} & \multicolumn{2}{c|}{FilterFL (Ours)} & \multicolumn{2}{c|}{No Defense} & \multicolumn{2}{c|}{FilterFL (Ours)} & \multicolumn{2}{c|}{No Defense} & \multicolumn{2}{c|}{FilterFL (Ours)} & \multicolumn{2}{c|}{No Defense} & \multicolumn{2}{c|}{FilterFL (Ours)} & \multicolumn{2}{c|}{No Defense} & \multicolumn{2}{c}{FilterFL (Ours)} \\ \cmidrule(l){2-21} 
 & MA(\%) & \multicolumn{1}{c|}{ASR(\%)} & MA(\%) & ASR(\%) & MA(\%) & \multicolumn{1}{c|}{ASR(\%)} & MA(\%) & ASR(\%) & MA(\%) & \multicolumn{1}{c|}{ASR(\%)} & MA(\%) & ASR(\%) & MA(\%) & \multicolumn{1}{c|}{ASR(\%)} & MA(\%) & ASR(\%) & MA(\%) & \multicolumn{1}{c|}{ASR(\%)} & MA(\%) & ASR(\%) \\ \midrule
\rowcolor{gray!15} BadNets (M) & 98.46 & \multicolumn{1}{c|}{99.83} & 98.55 & 0.31 & 84.92 & \multicolumn{1}{c|}{92.80} & 85.46 & 2.48 & 56.36 & \multicolumn{1}{c|}{99.52} & 54.95 & 0.65 & 95.51 & \multicolumn{1}{c|}{99.60} & 95.89 & 0.14 & 36.48 & \multicolumn{1}{c|}{99.89} & 36.32 & 0.57 \\
BadNets (S) & 96.39 & \multicolumn{1}{c|}{95.62} & 98.41 & 0.26 & 81.00 & \multicolumn{1}{c|}{67.65} & 85.89 & 1.89 & 52.15 & \multicolumn{1}{c|}{63.17} & 54.35 & 0.48 & 87.34 & \multicolumn{1}{c|}{68.56} & 95.72 & 0.15 & 35.02 & \multicolumn{1}{c|}{98.34} & 36.47 & 0.35 \\
\rowcolor{gray!15} Blended (M) & 98.25 & \multicolumn{1}{c|}{99.98} & 98.30 & 0.33 & 81.34 & \multicolumn{1}{c|}{93.43} & 82.19 & 3.08 & 55.96 & \multicolumn{1}{c|}{99.12} & 54.50 & 0.66 & 91.35 & \multicolumn{1}{c|}{99.31} & 91.19 & 0.41 & 36.25 & \multicolumn{1}{c|}{99.91} & 36.64 & 0.69 \\
Blended (S) & 97.25 & \multicolumn{1}{c|}{97.82} & 98.32 & 0.21 & 76.32 & \multicolumn{1}{c|}{71.28} & 85.62 & 1.93 & 50.50 & \multicolumn{1}{c|}{63.17} & 54.25 & 0.52 & 84.69 & \multicolumn{1}{c|}{69.16} & 95.36 & 0.16 & 35.22 & \multicolumn{1}{c|}{95.35} & 36.32 & 0.61 \\
\rowcolor{gray!15} Reflect (M) & 98.15 & \multicolumn{1}{c|}{95.43} & 98.34 & 0.35 & 83.22 & \multicolumn{1}{c|}{71.32} & 84.55 & 3.23 & 54.61 & \multicolumn{1}{c|}{97.73} & 54.32 & 0.59 & 93.36 & \multicolumn{1}{c|}{87.69} & 95.04 & 0.53 & 35.67 & \multicolumn{1}{c|}{99.53} & 36.52 & 0.95 \\
Reflect (S) & 95.98 & \multicolumn{1}{c|}{97.87} & 98.29 & 0.23 & 80.09 & \multicolumn{1}{c|}{52.10} & 85.43 & 1.62 & 51.37 & \multicolumn{1}{c|}{60.28} & 54.28 & 0.50 & 85.12 & \multicolumn{1}{c|}{62.30} & 95.48 & 0.13 & 34.83 & \multicolumn{1}{c|}{92.16} & 36.59 & 0.53 \\
\rowcolor{gray!15} Clean-Label (M) & 98.44 & \multicolumn{1}{c|}{57.12} & 98.26 & 0.27 & 84.25 & \multicolumn{1}{c|}{23.62} & 84.96 & 2.98 & 54.91 & \multicolumn{1}{c|}{25.62} & 54.45 & 0.61 & 94.37 & \multicolumn{1}{c|}{42.78} & 94.71 & 0.21 & 36.02 & \multicolumn{1}{c|}{82.37} & 36.28 & 0.58 \\
Clean-Label (S) & 96.12 & \multicolumn{1}{c|}{52.15} & 98.33 & 0.25 & 79.31 & \multicolumn{1}{c|}{20.08} & 85.57 & 1.51 & 51.32 & \multicolumn{1}{c|}{27.78} & 54.32 & 0.54 & 85.63 & \multicolumn{1}{c|}{25.12} & 95.37 & 0.14 & 35.10 & \multicolumn{1}{c|}{84.38} & 36.33 & 0.46 \\
\rowcolor{gray!15} DBA (M) & 98.32 & \multicolumn{1}{c|}{99.95} & 98.35 & 0.33 & 84.56 & \multicolumn{1}{c|}{97.61} & 84.93 & 2.34 & 55.37 & \multicolumn{1}{c|}{98.45} & 54.37 & 0.52 & 95.48 & \multicolumn{1}{c|}{99.99} & 95.56 & 0.23 & 36.44 & \multicolumn{1}{c|}{99.96} & 36.30 & 0.52 \\
DBA (S) & 95.87 & \multicolumn{1}{c|}{97.78} & 98.35 & 0.22 & 73.96 & \multicolumn{1}{c|}{79.75} & 86.15 & 2.04 & 50.56 & \multicolumn{1}{c|}{81.37} & 54.28 & 0.64 & 81.60 & \multicolumn{1}{c|}{81.35} & 95.42 & 0.17 & 35.36 & \multicolumn{1}{c|}{99.90} & 36.39 & 0.41 \\
\rowcolor{gray!15} Neurotoxin (M) & 97.57 & \multicolumn{1}{c|}{99.17} & 98.38 & 0.27 & 82.11 & \multicolumn{1}{c|}{93.13} & 85.42 & 1.72 & 54.76 & \multicolumn{1}{c|}{97.67} & 54.26 & 0.41 & 94.36 & \multicolumn{1}{c|}{99.92} & 95.18 & 0.18 & 36.48 & \multicolumn{1}{c|}{99.92} & 36.41 & 0.63 \\
Neurotoxin (S) & 97.12 & \multicolumn{1}{c|}{98.95} & 98.30 & 0.22 & 81.68 & \multicolumn{1}{c|}{88.63} & 85.63 & 1.65 & 51.25 & \multicolumn{1}{c|}{92.16} & 54.39 & 0.57 & 88.78 & \multicolumn{1}{c|}{92.42} & 95.67 & 0.21 & 36.05 & \multicolumn{1}{c|}{99.17} & 36.52 & 0.47 \\
\rowcolor{gray!15} A3FL & 98.25 & \multicolumn{1}{c|}{99.10} & 98.34 & 0.26 & 83.78 & \multicolumn{1}{c|}{96.33} & 84.62 & 2.92 & 52.83 & \multicolumn{1}{c|}{97.35} & 54.40 & 0.43 & 91.42 & \multicolumn{1}{c|}{94.22} & 95.36 & 0.30 &36.42  & \multicolumn{1}{c|}{99.03} & 36.41 & 0.50 \\
F3BA & 98.02 & \multicolumn{1}{c|}{98.32} & 98.23 & 0.31 & 81.17 & \multicolumn{1}{c|}{67.52} & 84.98 & 2.68 & 50.95 & \multicolumn{1}{c|}{82.16} & 54.28 & 0.62 & 88.26 & \multicolumn{1}{c|}{61.53} & 95.10 & 0.17 & 36.23 & \multicolumn{1}{c|}{99.30} & 36.37 & 0.42 \\ \bottomrule
\end{tabular}
}
\vspace{-0.15in}
\end{table*}

%

\subsection{Defense Adaptability Evaluation}\label{sec: adaptability}

{\bf  Performance with Different Sampling Rates.} 
To evaluate the defense performance of our approach in FL scenarios with different sampling rates, we conducted experiments on a system with 100 clients, where 10 of them are malicious. 
Specifically, we randomly selected $5\%$, $10\%$, $50\%$, and $100\%$ of the clients to participate in local training in each round, respectively. 
%
Note that when the sampling rate is $100\%$, all malicious clients must be present in every communication round.
Table~\ref{table: sampling rate} shows the experimental results on the CIFAR-10 dataset considering various IID and non-IID data distributions.
For each case in the experiments, we performed FL training until the global model converged.
From this table, we can see that FilterFL can achieve satisfactory defense performance in FL scenarios with different sampling rates. Note that even in the case with a $100\%$ sampling rate, where all malicious clients are present in each communication round, FilterFL can still effectively protect FL by detecting and excluding such malicious models.
%

{\bf  Performance against SOTA FL Backdoor Attacks.} 
%
To evaluate the defense performance of our approach against SOTA FL backdoor attacks, 
Table~\ref{table: attack} shows the defense performance of FilterFL on five datasets (i.e., MNIST, CIFAR-10, CIFAR-100, GTSRB, and Tiny-ImageNet) within a non-IID scenario ($\alpha=0.5$).
For BadNets~\cite{gu2017badnets}, Blended~\cite{chen2017targeted}, Reflect~\cite{liu2020reflection}, Clean-Label~\cite{turner2019label}, DBA~\cite{xie2020dba}, and Neurotoxin~\cite{zhang2022neurotoxin}, we considered their variants of multiple attacks (denoted by M) and single attacks (denoted by S),
respectively. 
%
For BadNets, which adds strip triggers at the corners of samples, we used the default trigger pattern mentioned in Section~\ref{sec: setup}.
For Blended, which blends triggers with entire samples,  we poisoned samples by blending them with trigger patterns in the form of a ``Hello Kitty'' image (with a transparency of 0.1).  
%
For Reflect, which uses invisible reflection triggers, we generated one specular reflection trigger for each sample following its original settings in \cite{liu2020reflection}. 
For Clean Label, which maintains the poisoned labels unchanged, we utilized our default trigger (i.e., a white stripe in the top left corner of the samples) and added adversarial perturbation to the samples in the same way as \cite{turner2019label}.
For DBA, it allows different malicious clients to use different parts of triggers to achieve a distributed backdoor attack, where we set the number of distributed triggers to 4.
For Neurotoxin, which aims to achieve durable attacks in FL, poisoned models prioritize updating parameters that are less likely to be updated in benign models, where we set the mask ratio to 0.25.
Note that all the multiple attacks in our experiments follow the default settings as described in Section~\ref{sec: setup}.
For each attack in our experiments, the malicious client uploaded its poisoned models to the server only during the last $10\%$ FL communication rounds, where the parameter weights of uploaded models are scaled up by a factor of 10. 
%
%
Based on their original settings (i.e., the number of malicious clients and the attack interval),  A3FL~\cite{ZhangJCLW23} and F3BA~\cite{FangC23} are two SOTA FL backdoor attacks that optimize triggers for each client according to the update trend of global models, aiming to extend the effective period of attacks.
From this table, we can find that FilterFL can effectively defend against the various complex attacks, showing the defense adaptability of our approach.

\begin{table}[ht]
\vspace{-0.1in}
\centering
\caption{Performance of FilterFL against complex triggers.}
\vspace{-0.1in}
\label{table: trigger}
\addtolength{\tabcolsep}{-3.5pt}
\scalebox{0.83}{
\begin{tabular}{c|cccc|cccc}
\toprule
\multirow{4}{*}{Trigger} & \multicolumn{4}{c|}{CIFAR-10} & \multicolumn{4}{c}{GTSRB} \\ \cmidrule(l){2-9} 
 & \multicolumn{2}{c|}{No Defense} & \multicolumn{2}{c|}{FilterFL (Ours)} & \multicolumn{2}{c|}{No Defense} & \multicolumn{2}{c}{FilterFL (Ours)} \\ \cmidrule(l){2-9} 
 & MA(\%) & \multicolumn{1}{c|}{ASR(\%)} & MA(\%) & ASR(\%) & MA(\%) & \multicolumn{1}{c|}{ASR(\%)} & MA(\%) & ASR(\%)\\ \midrule
\rowcolor{gray!15}Pattern 1 & 84.92 & \multicolumn{1}{c|}{92.80} & 85.46 & 2.48 & 95.51 & \multicolumn{1}{c|}{99.60} & 95.89 & 0.14 \\
Pattern 2  & 84.78 & \multicolumn{1}{c|}{62.23} & 85.33 & 2.86 & 95.63 & \multicolumn{1}{c|}{91.25} & 95.18 & 0.28 \\
\rowcolor{gray!15}Pattern 3  & 85.03 & \multicolumn{1}{c|}{86.57} & 85.27 & 2.23 & 94.50 & \multicolumn{1}{c|}{95.46} & 94.75 & 0.16 \\
Pattern 4  & 81.15 & \multicolumn{1}{c|}{12.33} & 80.12 & 2.52 & 91.00 & \multicolumn{1}{c|}{39.28} & 91.12 & 0.15 \\ \bottomrule
\end{tabular}
}
\vspace{-0.1in}
\end{table}

{\bf Performance against Complex Triggers.}
We conducted experiments to evaluate the defense performance of FilterFL against backdoor attacks with various more complex triggers in addition to the default trigger. 
Table~\ref{table: trigger} shows the experimental results for the CIFAR-10 and GTSRB datasets within a non-IID scenario ($\alpha=0.5$), where we used multiple attacks to perform backdoor attacks.
We followed the same settings in BadNets~\cite{gu2017badnets}, and considered four different trigger patterns (i.e., Patterns 1-4). Specifically, 
 Pattern 1 denotes the default trigger used in our experiments, i.e., a white stripe located in the upper left corner of the samples. 
 Pattern 2 indicates the case where the default triggers are distributed in all four corners of the samples.
 Patterns 3 and 4 maintain the same shape as patterns 1 and 2, respectively, but the color is red rather than white.
%
From this table, we can observe that FilterFL can defend against all the backdoor attacks, demonstrating the generalization ability of our approach in dealing with different complex triggers. 



\begin{table}[t]
\centering
\caption{Performance of FilterFL against multi-target attacks.}
\vspace{-0.1in}
\label{table: multitarget}
\addtolength{\tabcolsep}{-3pt}
\scalebox{0.82}{
\begin{tabular}{c|cccc|cccc}
\toprule
\multirow{4}{*}{\begin{tabular}[c]{c}Number\\ of \\ Targets\end{tabular}} & \multicolumn{4}{c|}{CIFAR-10} & \multicolumn{4}{c}{GTSRB} \\ \cmidrule(l){2-9} 
 & \multicolumn{2}{c|}{No Defense} & \multicolumn{2}{c|}{FilterFL (Ours)} & \multicolumn{2}{c|}{No Defense} & \multicolumn{2}{c}{FilterFL (Ours)} \\ \cmidrule(l){2-9} 
 & MA\% & \multicolumn{1}{c|}{ASR(\%)} & MA(\%) & ASR(\%) & MA(\%) & \multicolumn{1}{c|}{ASR(\%)} & MA(\%) & ASR(\%) \\ \midrule
\rowcolor{gray!15}1 & 84.92 & \multicolumn{1}{c|}{92.80} & 85.46 & 2.48 & 95.51 & \multicolumn{1}{c|}{99.60} & 95.45 & 0.14 \\
2 & 82.06 & \multicolumn{1}{c|}{77.23} & 84.37 & 2.46 & 93.43 & \multicolumn{1}{c|}{98.95} & 95.17 & 0.10 \\
\rowcolor{gray!15}3 & 80.15 & \multicolumn{1}{c|}{75.33} & 83.98 & 2.49 & 88.26 & \multicolumn{1}{c|}{96.23} & 94.72 & 0.18 \\
5 & 71.25 & \multicolumn{1}{c|}{54.49} & 82.56 & 2.12 & 85.38 & \multicolumn{1}{c|}{94.12} & 94.21 & 0.15 \\ \bottomrule
\end{tabular}
}
\vspace{-0.18in}
\end{table}
{\bf Performance against Multi-target Attacks.}
In real-world scenarios, adversaries may choose multiple categories as attack targets. 
Therefore, we conducted experiments to evaluate the defense performance of FilterFL against such multi-target backdoor attacks.
Here, we assume that adversaries design one specific trigger pattern for each target category. 
Table~\ref{table: multitarget} presents the experimental results on two datasets (i.e., CIFAR-10 and GTSRB) within a non-IID scenario ($\alpha=0.5$). 
As shown in the first column, we considered the defenses against a single-target backdoor attack and three multi-target attacks with 2, 3, and 5 attack targets, respectively.
From this table, we can observe that FilterFL can successfully defend against all attacks with low ASR. 
Interestingly, we also find that the MA is improved in seven out of eight cases, demonstrating the efficacy of our approach in FL-oriented backdoor defenses. For example, when the number of targets is $5$, FilterFL can improve MA by $8.83\%$ for the GTSRB dataset.


\begin{table}[ht]
\centering
\vspace{-0.1in}
\caption{Performance of FilterFL against adaptive attacks.}
\vspace{-0.1in}
\label{table: adaptive}
\addtolength{\tabcolsep}{-3.5pt}
\scalebox{0.82}{
\begin{tabular}{c|cccc|cccc}
\toprule
\multirow{4}{*}{\begin{tabular}[c]{c}Poisoning \\ Rate\end{tabular}} & \multicolumn{4}{c|}{CIFAR-10} & \multicolumn{4}{c}{GTSRB} \\ \cmidrule(l){2-9} 
 & \multicolumn{2}{c|}{No Defense} & \multicolumn{2}{c|}{FilterFL (Ours)} & \multicolumn{2}{c|}{No Defense} & \multicolumn{2}{c}{FilterFL (Ours)} \\ \cmidrule(l){2-9} 
 & MA(\%) & \multicolumn{1}{c|}{ASR(\%)} & MA(\%) & ASR(\%) & MA(\%) & \multicolumn{1}{c|}{ASR(\%)} & MA(\%) & ASR(\%) \\ \midrule
\rowcolor{gray!15}0.05 & 84.97 & \multicolumn{1}{c|}{45.37} & 85.43 & 2.27 & 95.44 & \multicolumn{1}{c|}{64.56} & 95.38 & 0.14 \\
0.10 & 85.02 & \multicolumn{1}{c|}{62.56} & 85.65 & 2.31 & 95.46 & \multicolumn{1}{c|}{90.23} & 95.41 & 0.16 \\
\rowcolor{gray!15}0.20 & 84.85 & \multicolumn{1}{c|}{81.42} & 85.51 & 2.15 & 95.38 & \multicolumn{1}{c|}{98.45} & 95.32 & 0.13 \\
0.30 & 84.92 & \multicolumn{1}{c|}{92.80} & 85.46 & 2.48 & 95.51 & \multicolumn{1}{c|}{99.60} & 95.45 & 0.14 \\ \bottomrule
\end{tabular}
}
\vspace{-0.1in}
\end{table}

{\bf Performance against Adaptive Attacks.} 
Assume that adversaries have all prior knowledge of our proposed approach to design an adaptive attack to bypass our defense approach. Based on our observation that backdoor knowledge can be learned more easily than normal knowledge,  adversaries can adaptively reduce the poisoning rates of their local data to slow down the processes of injecting backdoors into their local models. 
In this way, a backdoor can be slowly injected into global models over multiple rounds.
To evaluate the defense performance of FilterFL against adaptive attacks, we conducted experiments to apply attacks with different poisoning rates on the CIFAR-10 and GTSRB datasets, respectively,  within a non-IID setting (i.e., $\alpha=0.5$). Table~\ref{table: adaptive} shows the experimental results. 
From this table, we can observe that in the case of ``No Defense'' lower poisoning rates generally lead to lower ASR, indicating the difficulty for local models to learn backdoor knowledge when the poisoning rate is low.  However, FilterFL can effectively defend against backdoor attacks even when the poisoning rates are low, demonstrating that the adaptive attack with multiple rounds of injection can hardly bypass our proposed defense.

\subsection{Ablation Studies}
\begin{table}[ht]
\centering
\caption{Performance of FilterFL on different models.}
\vspace{-0.1in}
\label{table: model}
\addtolength{\tabcolsep}{-4.5pt}
\scalebox{0.80}{
\begin{tabular}{c|c|cc|cc|cc|cc}
\toprule
\multirow{2.5}{*}{Model} & \multirow{2.5}{*}{Dataset} & \multicolumn{2}{c|}{IID} & \multicolumn{2}{c|}{$\alpha = 1.0$} & \multicolumn{2}{c|}{$\alpha = 0.5$} & \multicolumn{2}{c}{$\alpha = 0.1$} \\ \cmidrule(l){3-10} 
 &  & MA(\%) & ASR(\%) & MA(\%) & ASR(\%) & MA(\%) & ASR(\%) & MA(\%) & ASR(\%) \\ \midrule  
  \rowcolor{gray!15}{\cellcolor{white}}& CIFAR-10 & 92.12 & 2.06 & 92.03 & 1.93 & 91.51 & 2.12 & 84.38 & 5.85 \\
ResNet & CIFAR-100 & 52.22 & 0.70 & 54.56 & 0.59 & 54.95 & 0.65 & 50.70 & 0.60\\
\rowcolor{gray!15}{\cellcolor{white}-18} & GTSRB & 98.51 & 0.12 & 97.62 & 0.15 & 97.33 & 0.14 & 96.23 & 0.48 \\ 
& T-ImgNet & 37.13 & 0.44 & 36.92 & 0.63 & 35.32 & 0.57 & 31.20 & 0.83 \\  \midrule
 \rowcolor{gray!15}{\cellcolor{white}}& CIFAR-10 & 94.15 & 1.97 & 94.07 & 1.92 & 93.27 & 2.12 & 86.45 & 5.23 \\
ResNet & CIFAR-100 & 62.16 & 0.61 & 62.83 & 0.62 & 61.29 & 0.51 & 60.62 & 0.53 \\
\rowcolor{gray!15}{\cellcolor{white}-34} & GTSRB & 98.62 & 0.13 & 97.96 & 0.12 & 97.37 & 0.13 & 97.15 & 0.45 \\
 & T-ImgNet & 51.93 & 0.47 & 51.36 & 0.62 & 49.82 & 0.52 & 45.14 & 1.01 \\ \midrule
 \rowcolor{gray!15}{\cellcolor{white}}& CIFAR-10 & 88.75 & 2.16 & 88.19 & 2.08 & 87.11 & 2.05 & 79.32 & 6.22 \\
VGG & CIFAR-100 & 55.38 & 0.74 & 56.92 & 0.60 & 55.49 & 0.57 & 52.77 & 0.65 \\
\rowcolor{gray!15}{\cellcolor{white}-16} & GTSRB & 97.42 & 0.10 & 96.71 & 0.12 & 96.54 & 0.11 & 96.23 & 0.46 \\
 & T-ImgNet & 42.56 & 0.42 & 42.45 & 0.61 & 41.97 & 0.43 & 37.28 & 0.98 \\ \bottomrule
\end{tabular}
}
\vspace{-0.1in}
\end{table}
{\bf Impact of Different Models.}
To evaluate the impacts of different models on FilterFL, we conducted experiments using different models as classifiers.
For CIFAR-10, CIFAR-100, GTSRB, and Tiny ImageNet datasets, we investigated three models (i.e., ResNet-18, ResNet-34~\cite{he2016identity}, and VGG-16~\cite{simonyan2014very}), respectively, considering different IID and non-IID data distributions. 
From Table~\ref{table: model}, we can find that FilterFL can achieve satisfactory defensive performance when adopting different models as classifiers, indicating that the effectiveness of FilterFL is independent of the underlying classifiers.



\begin{table}[t]
\centering
\caption{Ablation study on the second step (i.e., S2) considering different data distributions.}
\vspace{-0.1in}
\label{table: ablation}
\scalebox{0.85}{
\begin{tabular}{c|cc|cc|cc}
\toprule
\multicolumn{1}{c|}{\multirow{2.5}{*}{Distribution}} & \multicolumn{2}{c|}{No Defense} & \multicolumn{2}{c|}{FilterFL (Full)} & \multicolumn{2}{c}{FilterFL (w/o  S2)} \\ \cmidrule(l){2-7} 
\multicolumn{1}{c}{} & \multicolumn{1}{|c}{MA(\%)} & \multicolumn{1}{c|}{ASR(\%)} & \multicolumn{1}{c}{MA(\%)} & \multicolumn{1}{c|}{ASR(\%)} & \multicolumn{1}{c}{MA(\%)} & \multicolumn{1}{c}{ASR(\%)} \\ \midrule
\rowcolor{gray!15}IID & 84.33 & 91.73 & 86.66 & 2.33 & 72.32 & 2.65 \\
$\alpha=1.0$ & 85.22 & 88.92 & 86.70 & 2.27 & 73.51 & 10.32 \\
\rowcolor{gray!15}$\alpha=0.5$ & 84.92 & 92.80 & 85.46 & 2.48 & 72.57 & 41.61 \\
$\alpha=0.1$ & 77.12 & 97.04 & 72.63 & 7.42 & 67.35 & 89.69 \\ \bottomrule
\end{tabular}
}
\vspace{-0.1in}
\end{table}

\begin{table}[ht]
\centering
\caption{Ablation study on the second step (i.e., S2) disabled in different FL rounds for FilterFL.}
\vspace{-0.1in}
\label{table: ablation rounds}
\scalebox{0.85}{
\begin{tabular}{c|c|c|c|c|c|c}
\toprule
Start Round & 0 (``w/o S2'') & 400 & 800 & 1200 & 1600 & 2000 (Full) \\ \midrule
MA(\%) & 72.32 & 73.41 & 75.65 & 78.92 & 85.72 & 86.66 \\
ASR(\%) & 2.65 & 2.41 & 2.36 & 2.32 & 2.32 & 2.33 \\ \bottomrule
\end{tabular}
}
\vspace{-0.15in}
\end{table}

\begin{table*}[h]
\centering
\caption{Performance of FilterFL with different hyperparameter settings for $\rho$ on CIFAR-10. }
\vspace{-0.1in}
\label{table: rho}
\addtolength{\tabcolsep}{-2pt}
\scalebox{0.9}{
\begin{tabular}{@{}c|cccc|cccc|cccc|cccc@{}}
\toprule
 & \multicolumn{4}{c|}{IID} & \multicolumn{4}{c|}{$\alpha=1.0$} & \multicolumn{4}{c|}{$\alpha=0.5$} & \multicolumn{4}{c}{$\alpha=0.1$} \\ \cmidrule(l){2-17} 
\multirow{-2.5}{*}{$\rho$} & MA(\%) & \multicolumn{1}{c|}{ASR(\%)} & FRR(\%) & FAR(\%) & MA(\%) & \multicolumn{1}{c|}{ASR(\%)} & FRR(\%) & FAR(\%) & MA(\%) & \multicolumn{1}{c|}{ASR(\%)} & FRR(\%) & FAR(\%) & MA(\%) & ASR(\%) & FRR(\%) & FAR(\%) \\ \midrule
\rowcolor{gray!15}0.20 & 85.58 & \multicolumn{1}{c|}{2.25} & 4.23 & 2.50 & 81.51 & \multicolumn{1}{c|}{2.34} & 17.17 & 2.91 & 65.34 & \multicolumn{1}{c|}{3.63} & 32.38 & 3.43 & 45.63 & 8.31 & 39.21 & 4.14 \\
0.30 & 86.26 & \multicolumn{1}{c|}{2.40} & 4.09 & 2.87 & 86.51 & \multicolumn{1}{c|}{2.12} & 3.97 & 3.02 & 81.89 & \multicolumn{1}{c|}{2.57} & 13.46 & 3.38 & 62.45 & 7.66 & 15.76 & 4.31 \\
\rowcolor{gray!15}0.40 & 86.54 & \multicolumn{1}{c|}{2.26} & 3.58 & 3.05 & 86.69 & \multicolumn{1}{c|}{2.27} & 3.51 & 3.12 & 85.46 & \multicolumn{1}{c|}{2.44} & 6.12 & 3.39 & 72.59 & 7.44 & 7.33 & 6.17 \\
0.50 & 86.66 & \multicolumn{1}{c|}{2.33} & 3.15 & 3.22 & 86.70 & \multicolumn{1}{c|}{2.27} & 3.02 & 3.18 & 85.46 & \multicolumn{1}{c|}{2.48} & 3.17 & 3.42 & 72.63 & 7.42 & 4.03 & 8.02 \\
\rowcolor{gray!15}0.60 & 86.53 & \multicolumn{1}{c|}{2.31} & 3.10 & 3.25 & 86.66 & \multicolumn{1}{c|}{3.26} & 3.08 & 3.28 & 84.58 & \multicolumn{1}{c|}{15.87} & 3.08 & 12.38 & 75.45 & 45.52 & 3.97 & 38.83 \\
0.70 & 85.45 & \multicolumn{1}{c|}{7.46} & 3.16 & 3.64 & 86.57 & \multicolumn{1}{c|}{10.02} & 3.12 & 9.77 & 84.97 & \multicolumn{1}{c|}{62.34} & 3.12 & 28.34 & 76.69 & 99.23 & 3.38 & 64.16 \\ \bottomrule
\end{tabular}
}
\vspace{-0.1in}
\end{table*}

{\bf Impact of Trigger Filtering.}
As described in Section~\ref{section: Trigger Filtering}, the second step of our approach (i.e., trigger filtering) plays an important role in distinguishing between normal classification knowledge and backdoor knowledge. 
To demonstrate the effect of this step, we considered a variant (i.e., ``w/o S2'') of FilterFL by removing its step 2. Without changing other settings of FilterFL, this variant directly feeds the $\mathbf{S}_{g}$ generated by Step 1 into Step 3. 
We compared the defense performance between the variant and the full version of FilterFL, aiming to investigate the impact of trigger filtering. 
Table~\ref{table: ablation} presents the experimental results on the CIFAR-10 dataset, considering the impacts of different data distributions.  
%
%
From the table, we can find that within the IID scenario, although both the variant and the full version can achieve low ASR,  the variant fails to maintain high MA.
This is because the lack of Step 2 leads to the improper exclusion of benign models. 
%
%
%
For the non-IID cases, we can observe that the defense performance of both the variant and the full version deteriorates when the value of $\alpha$ decreases. 
Specifically, when $\alpha=0.1$, the variant becomes ineffective (ASR $=$ $89.69\%$) to defend against backdoor attacks, while the defense performance of the full version is tolerable (ASR $=$ $7.42\%$). 
Here, we find that the full version outperforms the variant dramatically in terms of both MA and ASR, showing the importance of the second step in backdoor defense. 
%

We also investigated the impact of trigger filtering for FL training processes. 
Table~\ref{table: ablation rounds} presents the results of the ablation study on the CIFAR-10 dataset within an IID scenario, where the second step of FilterFL is disabled starting from five different FL rounds (i.e., 0, 400, 800, 1200, 1600, and 2000). Here, we assume that backdoor attacks are launched from round 0 for all five cases.
Since each FL training on CIFAR-10 involves $2000$ communication rounds in total, if the disablement of Step 2 starts from round 2000, the final defense performance will equal that of the full version of FilterFL.
In contrast, if the disablement starts from round 0, the final defense performance will be the same as that of the variant ``w/o S2''.   
From this table, we can find that the earlier the disablement is performed, the worse the final defense performance. Among the five cases, the full version achieves the best defense performance. Note that although all five cases can achieve tolerable ASR, their MA differences are significant. This is because, without applying Step 2, FilterFL will generate more false positives in detecting poisoned models, thus lowering the overall MA. 

\begin{table*}[t]
\centering
\caption{Ablation study on different numbers of training rounds for CGAN-1 (i.e., $E_1$). }
\vspace{-0.1in}
\label{table: cgan1}
\addtolength{\tabcolsep}{-3.2pt}
\scalebox{0.92}{
\begin{tabular}{c|cc|cc|cc|cc|cc|cc|cc|cc}
\toprule
\multirow{2.5}{*}{Distribution} & \multicolumn{2}{c|}{$E_1=2$} & \multicolumn{2}{c|}{$E_1=6$} & \multicolumn{2}{c|}{$E_1=8$} & \multicolumn{2}{c|}{$E_1=10$} & \multicolumn{2}{c|}{$E_1=12$} & \multicolumn{2}{c|}{$E_1=16$} & \multicolumn{2}{c|}{$E_1=20$} & \multicolumn{2}{c}{$E_1=24$} \\ \cmidrule(l){2-17} 
 & MA(\%) & ASR(\%) & MA(\%) & ASR(\%) & MA(\%) & ASR(\%) & MA(\%) & ASR(\%) & MA(\%) & ASR(\%) & MA(\%) & ASR(\%) & MA(\%) & ASR(\%)& MA(\%) & ASR(\%) \\ \midrule
\rowcolor{gray!15}IID & 55.91 & 24.23 & 54.25 & 0.75 & 54.23 & 0.70 & 54.25& 0.74& 54.22 & 0.73 & 54.23 & 0.72 & 54.25 & 0.74 &54.24 &0.73\\
$\alpha=1.0$ & 56.07 & 26.98 & 54.98 & 7.59 & 54.61 & 0.61 & 54.57& 0.52& 54.56 & 0.53 & 54.55 & 0.56 & 54.53 & 0.51 &54.55 &0.52\\
\rowcolor{gray!15} $\alpha=0.5$ & 55.89 & 59.28 & 55.33 & 22.06 & 55.01 & 4.98 &54.98 & 0.72& 54.95 & 0.65 & 54.98 & 0.63 & 54.93 & 0.59 & 54.96 & 0.60\\
$\alpha=0.1$ & 54.91 & 87.63 & 53.43 & 30.88 & 51.78 & 3.21 &51.15 &0.92 & 50.74 & 0.64 & 50.71 & 0.59 & 50.72 & 0.61 &50.74 &0.63\\ \bottomrule
\end{tabular}
\vspace{-0.3in}
}
\end{table*}


{\bf Impact of Hyperparameter $\rho$.}
To understand the impact of the hyperparameter $\rho$ on FilterFL, we conducted experiments considering different settings for $\rho$.  Due to the limited space, Table~\ref{table: rho} only presents the experimental results on the CIFAR-10 dataset. Note that FilterFL shows similar defense performance on the other datasets.
We can find that, for a fixed $\rho$,  the more non-IID the local data, the lower MA and higher ASR we can achieve. 
Moreover, we can observe that when dealing with a non-IID setting (e.g., $\alpha =0.5$ or $\alpha =0.1$), both MA and ASR decrease significantly as $\rho$ decreases.
To understand why the MA and ASR of FilterFL change significantly for different $\rho$, we conducted experiments to investigate two metrics: i) the False Rejection Rate (FRR), indicating the probability that a benign model is identified as a poisoned one; and ii) the False Acceptance Rate (FAR), denoting the probability that a poisoned model is identified as a
benign one.
%
From Table~\ref{table: rho}, we observe that when $\rho>0.3$, FilterFL can achieve an acceptable FRR, meaning that our approach will not exclude too many benign models, which hardly reduces the MA.
Meanwhile, we find that, along with the increase in $\rho$, FRR decreases and FAR increases.
It means that, although a smaller $\rho$ can effectively exclude more poisoned models from participating in aggregation, it also eliminates more benign ones, leading to a lower MA. 
Note that when $\rho$ becomes larger,  FilterFL will result in a worse ASR for the global model, since more poisoned models are neglected during aggregation. 
For example, when $\alpha=0.1$, we can observe the strong impact of $\rho$ on the defense performance of FilterFL.  
%
In this paper, we assume that the defenders have no prior knowledge of the degree of data heterogeneity. To properly make a trade-off between ASR and MA in defense, according to the results in Table~\ref{table: rho}, we suggest setting $\rho$ to 0.5 by default in FilterFL. 
In this paper, we set $\rho=0.5$ for all experiments conducted on the GTSRB, CIFAR-10, CIFAR-100, and Tiny-ImageNet datasets. 
Since MNIST samples are single channel and small in size, making knowledge extraction by FilterFL easier compared with other datasets, we set $\rho$ to 0.3 for the MNIST dataset.
Note that FilterFL used the fixed hyperparameters in extensive experiments involving four IID and non-IID data distributions, three model architectures, and eight backdoor attacks. This indicates that our approach does not require the specific adjustment of hyperparameters for different scenarios, which shows the applicability of our approach.

\begin{table*}[t]
\centering
\caption{Ablation study on different numbers of training rounds for CGAN-2 (i.e., $E_2$). 
}
\vspace{-0.1in}
\label{table: cgan2}
\addtolength{\tabcolsep}{-3.2pt}
\scalebox{0.92}{
\begin{tabular}{c|cc|cc|cc|cc|cc|cc|cc|cc}
\toprule
\multirow{2.5}{*}{Distribution} & \multicolumn{2}{c|}{$E_2=3$} & \multicolumn{2}{c|}{$E_2=6$} & \multicolumn{2}{c|}{$E_2=9$} & \multicolumn{2}{c|}{$E_2=12$} & \multicolumn{2}{c|}{$E_2=15$} & \multicolumn{2}{c|}{$E_2=20$} & \multicolumn{2}{c|}{$E_2=25$} & \multicolumn{2}{c}{$E_2=30$} \\ \cmidrule(l){2-17} 
& MA(\%) & ASR(\%) & MA(\%) & ASR(\%) & MA(\%) & ASR(\%) & MA(\%) & ASR(\%) & MA(\%) & ASR(\%) & MA(\%) & ASR(\%)& MA(\%) & ASR(\%)& MA(\%) & ASR(\%) \\ \midrule
\rowcolor{gray!15} IID & 54.21 & 1.23 & 54.45 & 0.75 & 54.55 & 0.72 & 54.62 & 0.73 & 54.63 & 0.72 & 54.61 & 0.74 & 54.62& 0.73&54.61 & 0.71\\
$\alpha=1.0$ & 53.07 & 26.98 & 54.08 & 7.59 & 54.61 & 0.61 & 54.56 & 0.53 & 54.55 & 0.56 & 54.53 & 0.51  & 54.50 & 0.55& 54.52&0.52\\
\rowcolor{gray!15} $\alpha=0.5$ & 52.89 & 69.28 & 54.33 & 22.06 & 55.01 & 4.98 & 54.95 & 0.95 & 54.95 & 0.63 & 54.93 & 0.59 & 55.01& 0.60 & 54.95&0.58\\
$\alpha=0.1$ & 48.62 & 89.62 & 49.31 & 42.88 & 49.78 & 13.21 & 50.12 & 1.73 & 50.71 & 0.59 & 50.70 & 0.61  & 50.73& 0.55& 50.69&0.61\\ \bottomrule
\end{tabular}
}
\vspace{-0.15in}
\end{table*}


{\bf Impact of the Numbers of CGAN Training Rounds.}
%
We conducted experiments to investigate the impact of the number of training rounds for CGAN-1 and CGAN-2, denoted by $E_1$ and $E_2$, respectively. 
Table~\ref{table: cgan1} and Table~\ref{table: cgan2} show the defense performance results of FilterFL on CIFAR-100. 
Specifically, in Table~\ref{table: cgan1}, we fixed the value of $E_2$ (i.e., $E_2=15$), while in Table~\ref{table: cgan2}, we fixed the value of $E_1$ (i.e., $E_1 = 12$). 
%
%
From these two tables, we can find that the defense performance of FilterFL becomes better when $E_1$ or $E_2$ increases, especially when $\alpha=0.1$. This means that without sufficient training on the underlying CGANs, the defense performance of FilterFL cannot be guaranteed. 
Specifically, Table~\ref{table: cgan1} shows that with an increase of $E_1$, MA drops slightly, but ASR decreases rapidly regardless of the data distributions. 
This is because, with fewer training rounds (i.e., with a small $E_1$), CGAN-1 cannot extract sufficient backdoor knowledge, thus hindering the correct exclusion of poisoned models.
%
%
Meanwhile, Table~\ref{table: cgan2} shows that as $E_2$ increases, MA increases slightly while ASR decreases sharply (when $E_2 < 15$) for different data distributions. It means that, without sufficient training, it is difficult for CGAN-2 to distinguish trigger knowledge from normal knowledge accurately, resulting in the unexpected exclusion of benign models for aggregation. 
Although increasing the number of training rounds of CGAN-1 and CGAN-2 can improve overall defense performance, it inevitably results in extra time overhead (see Section~\ref{section: discussion} for more details), which slows down the overall training process.  
Based on the experimental results shown in Table~\ref{table: cgan1} and Table~\ref{table: cgan2}, we make a proper trade-off between defense performance and training time of CGAN-1 and CGAN-2, where we set the number of training rounds for CGAN-1 and CGAN-2 at 12 and 15, respectively, by default.

\vspace{-0.05in}
\subsection{Discussion}\label{section: discussion}
{\bf Privacy Protection.}
Since FilterFL leverages CGANs to extract implicit backdoor knowledge rather than to restore original training samples, the extracted knowledge by CGANs does not involve any semantic information about the training samples. In other words, FilterFL does not compromise the privacy of user data. 
To verify this claim, we conducted an experiment on CIFAR-10 within a non-IID scenario ($\alpha=0.5$). Figure~\ref{fig: extracted} (in the Supplementary Material) shows the generated samples by CGANs from different training rounds, which do not leak any private information of the original training data.
Additionally, we conducted another experiment to check whether FilterFL is compatible with existing privacy protection methods for FL.
Since  Local Differential Privacy (LDP)~\cite{arachchige2019local,zhao2020local} is recognized as a promising means to prevent privacy leakage of clients and defend against inference attacks~\cite{nguyen2019diot,ganju2018property,shokri2017membership}, we adopted it as the underlying privacy protection mechanism of FilterFL. 
To evaluate the compatibility between FilterFL and LDP, we conducted an experiment on the MNIST dataset within a non-IID scenario ($\alpha=0.5$), where clients used LDP to protect their local data. 
Specifically, we considered two scenarios: i) benign clients use LDP to protect their private data while malicious clients do not, and ii) both benign and malicious clients use LDP. 
We use $\varepsilon$ to specify the LDP privacy budget, where a smaller value of $\varepsilon$ indicates a higher level of protection.

\begin{table}[ht]
\centering
\caption{Defense performance on MNIST under privacy protection within a non-IID scenario ($\alpha=0.5$).}
\vspace{-0.1in}
\label{table: privacy}
\addtolength{\tabcolsep}{-4.5pt}
\scalebox{0.83}{
\begin{tabular}{c|cc|cc|cc|cc|cc}
\toprule
\multirow{4}{*}{$\varepsilon$} & \multicolumn{2}{c|}{\multirow{2.5}{*}{No Attack}} & \multicolumn{4}{c|}{Malicious Clients w/o LDP} & \multicolumn{4}{c}{Malicious Clients w/ LDP} \\ \cmidrule(l){4-11} 
 & \multicolumn{2}{c|}{} & \multicolumn{2}{c|}{No Defense} & \multicolumn{2}{c|}{FilterFL (Ours)} & \multicolumn{2}{c|}{No Defense} & \multicolumn{2}{c}{FilterFL (Ours)} \\ \cmidrule(l){2-11} 
 & \multicolumn{1}{c}{MA(\%)} & \multicolumn{1}{c|}{ASR(\%)} & MA(\%) & \multicolumn{1}{c|}{ASR(\%)} & \multicolumn{1}{c}{MA(\%)} & ASR(\%) & MA(\%) & \multicolumn{1}{c|}{ASR(\%)} & \multicolumn{1}{c}{MA(\%)} & \multicolumn{1}{c}{ASR(\%)} \\ \midrule
\rowcolor{gray!15}$+\infty$ & 98.38 & 0.22 & 98.46 & 99.83 & 98.55 & 0.31 & 98.46 & 99.83 & 98.55 & 0.31 \\
30 & 92.70 & 0.42 & 91.43 & 99.77 & 91.23 & 0.33 & 91.23 & 99.88 & 92.31 & 0.37 \\
\rowcolor{gray!15}20 & 91.86 & 0.38 & 91.19 & 99.69 & 90.62 & 0.28 & 90.45 & 99.63 & 90.05 & 0.32 \\
10 & 87.57 & 0.56 & 87.02 & 99.89 & 85.56 & 0.42 & 85.76 & 96.24 & 84.33 & 0.33 \\
\rowcolor{gray!15}5 & 58.21 & 1.08 & 57.66 & 99.94 & 51.23 & 0.98 & 57.62 & 91.66 & 57.25 & 1.06 \\
1 & 19.27 & 1.32 & 19.22 & 99.91 & 18.96 & 1.23 & 15.43 & 85.47 & 15.34 & 1.35 \\ \bottomrule
\end{tabular}
}
\vspace{-0.1in}
\end{table}

Table~\ref{table: privacy} presents the evaluation results. Here, the third column indicates that LDP is applied only to benign models, and the fourth column denotes that LDP is applied to all the models. 
Note that for the case of $\varepsilon\neq +\infty$, LDP will inevitably deteriorate MA, especially when $\varepsilon$ is small.
For the case of $\varepsilon= +\infty$, LDP will not be applied to any models. 
From this table, we can observe that LDP-based vanilla FL (shown in the ``No Defense'' subcolumns of the third and fourth columns) fails to resist backdoor attacks, while the combination of
LDP and our approach can effectively address the backdoor attack problem, whether or not malicious clients use LDP.  
In other words, LDP can be easily integrated into FilterFL to enhance the privacy protection of local models, while the overall defense performance of FilterFL remains almost the same.

%


\begin{table}[ht]
\vspace{-0.1in}
\centering
\caption{Comparison of extra time overhead in one FL training round caused by different backdoor defense methods.}
\vspace{-0.1in}
\label{fig: overhead}
\addtolength{\tabcolsep}{-3.5pt}
\scalebox{0.85}{
\begin{tabular}{@{}c|ccc|ccc@{}}
\toprule
\multirow{2.5}{*}{Method} & \multicolumn{3}{c|}{CIFAR-10} & \multicolumn{3}{c}{CIFAR-100} \\ \cmidrule(l){2-7} 
 & \multicolumn{1}{c|}{Overhead (s)} & \multicolumn{1}{c|}{MA(\%)} & ASR(\%) & \multicolumn{1}{c|}{Overhead (s)} & \multicolumn{1}{c|}{MA(\%)} & ASR(\%) \\ \midrule
No Defense & \multicolumn{1}{c|}{18.462} & \multicolumn{1}{c|}{84.92} & 92.80 & \multicolumn{1}{c|}{31.283} & \multicolumn{1}{c|}{56.36} & 99.52 \\ \midrule
\rowcolor{gray!15}Krum & \multicolumn{1}{c|}{18.462 + 0.516} & \multicolumn{1}{c|}{70.66} & 7.22 & \multicolumn{1}{c|}{31.283 + 0.663} & \multicolumn{1}{c|}{41.27} & \textbf{0.64} \\
COMED & \multicolumn{1}{c|}{18.462 + 0.052} & \multicolumn{1}{c|}{84.08} & 61.30 & \multicolumn{1}{c|}{31.283 + 0.059} & \multicolumn{1}{c|}{51.81} & 99.43 \\
\rowcolor{gray!15}RLR & \multicolumn{1}{c|}{18.462 + 0.056} & \multicolumn{1}{c|}{76.93} & 8.30 & \multicolumn{1}{c|}{31.283 + 0.063} & \multicolumn{1}{c|}{26.64} & 42.25 \\
Trimmed-Mean & \multicolumn{1}{c|}{18.462 + 0.051} & \multicolumn{1}{c|}{84.59} & 64.54 & \multicolumn{1}{c|}{31.283 + 0.066} & \multicolumn{1}{c|}{52.04} & 99.47 \\
\rowcolor{gray!15}DP & \multicolumn{1}{c|}{18.462 + 0.054} & \multicolumn{1}{c|}{71.23} & 84.21 & \multicolumn{1}{c|}{31.283 + 0.055} & \multicolumn{1}{c|}{41.56} & 89.70 \\
FLAME & \multicolumn{1}{c|}{18.462 + 0.833} & \multicolumn{1}{c|}{81.85} & 93.78 & \multicolumn{1}{c|}{31.283 + 1.132} & \multicolumn{1}{c|}{51.52} & 98.24 \\
\rowcolor{gray!15}FLTrust & \multicolumn{1}{c|}{18.462 + 0.716} & \multicolumn{1}{c|}{82.55} & 20.53 & \multicolumn{1}{c|}{31.283 + 1.068} & \multicolumn{1}{c|}{53.45} & 29.69 \\
\rowcolor{gray!45}FilterFL (Ours) & \multicolumn{1}{c|}{18.462 + 0.836} & \multicolumn{1}{c|}{\textbf{85.46}} & \textbf{2.48} & \multicolumn{1}{c|}{31.283 + 1.826} & \multicolumn{1}{c|}{\textbf{54.95}} & 0.65 \\ \bottomrule
\end{tabular}
}
\vspace{-0.05in}
\end{table}


{\bf Extra Time Overhead.}
We conducted experiments to evaluate the extra time overhead caused by the two CGANs in FilterFL. Table~\ref{fig: overhead} compares the average extra time overhead in one FL round caused by different backdoor defense methods on CIFAR-10 and CIFAR-100 within a non-IID scenario ($\alpha=0.5$), and presents their final defense performance in round 2000.  
Unlike Table~\ref{table: datasets}, this table does not investigate FLIP~\cite{zhang2022flip} as it is deployed in clients rather than on the server, making the comparison unfair.  
In this table, we use ``No Defense'' to denote the case of vanilla FL and divide the time overhead of other defense methods into two parts: i) the training time of backbone FL (i.e., vanilla FL), and ii) extra time overhead caused by specific defense methods.  
From this table, we can find that, compared with SOTA defense methods, 
FilterFL can achieve much better defense performance in terms of MA and ASR with similar or tolerable extra time overhead. Note that, compared with the training time of the backbone FL in each round, the extra time overhead caused by FilterFL is negligible (e.g., the time overhead caused by CGANs is only $4.53\%$ of the FL training time).

%

%

\section{Conclusion}\label{section:6}

Although various solutions have been investigated to defend against Federated Learning (FL)-oriented backdoor attacks, most of them rely on additional auxiliary training data, greatly violating the data privacy policy imposed by FL. To address this problem, this paper first reveals the characteristic differences between normal knowledge and backdoor knowledge learned by the global model, which can be used to identify backdoor injection operations of malicious clients. Based on our proposed knowledge extraction and backdoor filtering schemes, we present a novel data-free backdoor defense method named FilterFL. In each FL communication round, FilterFL can effectively exclude backdoor knowledge contributed by malicious clients, thus maximizing overall inference performance while effectively declining backdoor injections into the global model.   
Comprehensive experimental results on well-known datasets show the effectiveness and applicability of our approach within various data heterogeneity scenarios. 

\begin{acks}
This work was supported by the Natural Science Foundation of China (62272170), ``Digital Silk Road'' Shanghai International Joint Lab of Trustworthy Intelligent Software (22510750100), Shanghai Trusted Industry Internet Software Collaborative Innovation Center, the National Research Foundation, Singapore, and Cyber Security Agency of Singapore under its National Cybersecurity R\&D Programme and CyberSG R\&D Cyber Research Programme Office. Any opinions, findings and conclusions or recommendations expressed in this material are those of the author(s) and do not reflect the views of National Research Foundation, Singapore, Cyber Security Agency of Singapore as well as CyberSG R\&D Programme Office, Singapore.
The authors Yanxin Yang, Pengyu Zhang, and Mingsong Chen are with the MoE Engineering Research Center of Hardware/Software Co-design Technology and Application at East China Normal University, Shanghai, China.
\end{acks}

\bibliographystyle{ACM-Reference-Format}
\bibliography{reference}

\appendix
\section*{Supplemental Materials}

\section{Additional Experimental Results}
\subsection{More Performance Comparisons}
We compared our approach with ten SOTA baselines on five well-known datasets (i.e., MNIST, CIFAR-10, CIFAR-100, GTSRB, and Tiny-ImageNet) within one IID scenario and three non-IID scenarios (with $\alpha=0.1$, $0.5$, and $1.0$). The experimental results are shown in Table~\ref{table: complete}. 
From the table, we can find that, compared with the baselines, our approach minimizes ASR while maintaining high MA on different datasets within varying data distribution scenarios.

\subsection{Examples of Generated Samples}
\begin{figure}[b]
\centering
\includegraphics[width=0.45\textwidth]{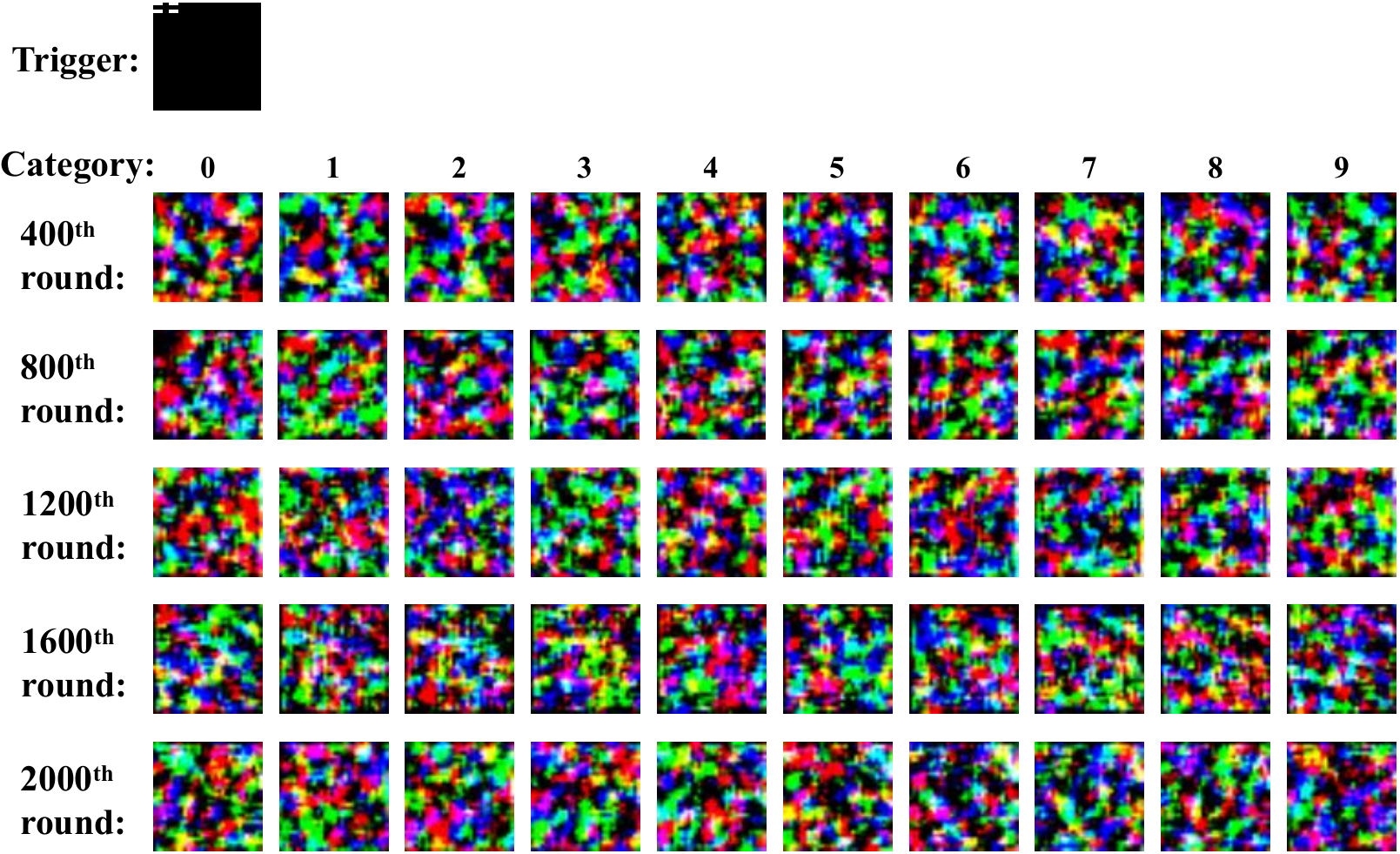}
\caption{Sample set $\mathbf{T}_g$ generated by CGAN-2 in different FL communication rounds.}
\label{fig: extracted}
\end{figure}

We conducted experiments on CIFAR-10 within a non-IID scenario ($\alpha=0.5$) and saved the sample set $\mathbf{T}_g$ generated by CGANs in different FL communication rounds, shown in Figure~\ref{fig: extracted}. 
From the figure, we can observe that the samples do not contain any semantic information about benign data, meaning that our approach does not cause any privacy leaks.
Meanwhile, we can find that the generated samples are not related in shape to the actual trigger. This is because FilterFL focuses on generating samples with the same backdoor function rather than recovering the actual triggers.

\begin{table}[b]
\centering
\caption{Performance comparison on five datasets within varying data distribution scenarios.}
\addtolength{\tabcolsep}{-1.5pt}
\scalebox{0.5}{
\label{table: complete}
\begin{tabular}{c|c|cc|cc|cc|cc|cc}
\toprule
\multirow{2.5}{*}{\begin{tabular}[c]{c}Data \\ Distribution\end{tabular}} & \multirow{2.5}{*}{Method} & \multicolumn{2}{c|}{MNIST} & \multicolumn{2}{c|}{CIFAR-10} & \multicolumn{2}{c|}{CIFAR-100} & \multicolumn{2}{c|}{GTSRB} & \multicolumn{2}{c}{Tiny-ImageNet} \\ \cmidrule(l){3-12} 
 &  & MA(\%) & ASR(\%) & MA(\%) & ASR(\%) & MA(\%) & ASR(\%) & MA(\%) & ASR(\%) & MA(\%) & ASR(\%) \\ \midrule
& No Attack & 98.70 & 0.12 & 86.53 & 2.31 & 52.28 & 0.60 & 96.52 & 0.10 & 37.31 & 0.30 \\
 & No Defense & 98.77 & 99.84 & 84.33 & 91.73 & 54.55 & 99.71 & 96.59 & 99.76 & 38.76 & 99.87 \\ \cmidrule(l){2-12} 
 \rowcolor{gray!15}{\cellcolor{white}}& Krum & 97.59 & 0.34 & 84.78 & 2.76 & 50.19 & \textbf{0.70} & 95.74 & 0.17 & 35.17 & 0.93 \\
 & COMED & 98.50 & 0.43 & 86.28 & 10.76 & 53.80 & 99.39 & 96.26 & 91.29 & 36.39 & 96.52 \\
 \rowcolor{gray!15}{\cellcolor{white}}& RLR & 98.40 & 61.94 & 78.84 & 3.89 & 35.78 & 23.44 & 94.88 & 0.67 & 31.29 & 12.74 \\
 & Trimmed Mean & 98.00 & 1.13 & 86.20 & 14.32 & \textbf{53.88} & 99.40 & 96.21 & 93.92 & 36.32 & 95.88 \\
 \rowcolor{gray!15}{\cellcolor{white}}& DP & 90.32 & 84.25 & 71.13 & 84.48 & 38.65 & 89.86 & 91.95 & 88.73 & 28.56 & 81.07 \\
 & FLAME & 98.45 & 0.32 & 85.56 & 10.85 & 51.81 & 1.49 & 95.82 & 33.82 & 36.42 & 22.07 \\
 \rowcolor{gray!15}{\cellcolor{white}}& FLTrust & 98.38 & 0.29 & 84.42 & 2.88 & 51.34 & 0.85 & 96.05 & 0.21 & \textbf{37.13} & 0.57 \\
 & FLIP & 98.51 & 0.42 & 86.65 & 2.69 & 51.68 & 2.31 & 96.21 & 0.65 & 36.92 & 0.87 \\
 \rowcolor{gray!15}{\cellcolor{white}}& FLDetector & \textbf{98.61} & 0.35 & 86.23 & 2.71 & 50.39 & 0.98 & 96.09 & 0.25 & 35.20 & 28.04\\
 & FedREDefense  & 98.73 & 0.31 & 85.14 & 2.35 & 51.86 & \textbf{0.70} & 96.17 & 0.15 & 37.05 & 0.50\\
 \rowcolor{gray!45}{\cellcolor{white}\multirow{-12}{*}{IID}} & FilterFL (Ours) & 98.60 & \textbf{0.20} & \textbf{86.66} & \textbf{2.33} & 52.22 & \textbf{0.70} & \textbf{96.36} & \textbf{0.11} & \textbf{37.13} & \textbf{0.44} \\ \midrule
 & No Attack & 98.27 & 0.24 & 86.59 & 2.06 & 55.08 & 0.70 & 95.63 & 0.12 & 37.23 & 0.21 \\
 & No Defense & 98.75 & 99.57 & 85.22 & 88.92 & 56.49 & 99.58 & 95.89 & 99.75 & 38.88 & 99.69 \\ \cmidrule(l){2-12} 
 \rowcolor{gray!15}{\cellcolor{white}}& Krum & 95.84 & 0.49 & 77.83 & 3.26 & 50.31 & \textbf{0.42} & 95.23 & 0.29 & 30.27 & 0.96 \\
 & COMED & 98.48 & 0.44 & 85.95 & 30.22 & 54.30 & 99.37 & \textbf{95.74} & 90.46 & 35.43 & 99.87 \\
 \rowcolor{gray!15}{\cellcolor{white}}& RLR & 98.29 & 11.02 & 80.58 & \textbf{0.80} & 34.56 & 28.68 & 94.31 & 49.52 & 22.38 & 32.06 \\
 & Trimmed Mean & 98.50 & 0.72 & 86.10 & 30.50 & 54.49 & 99.50 & 91.76 & 94.50 & 34.88 & 99.67 \\
 \rowcolor{gray!15}{\cellcolor{white}}& DP & 91.61 & 86.87 & 71.95 & 84.05 & 41.33 & 89.66 & 90.69 & 88.56 & 27.64 & 78.38 \\
 & FLAME & 98.21 & 0.30 & 84.24 & 87.17 & 52.32 & 81.21 & 94.95 & 44.84 & 35.29 & 42.31 \\
 \rowcolor{gray!15}{\cellcolor{white}}& FLTrust & 98.36 & 0.33 & 84.56 & 3.83 & 53.55 & 2.78 & 95.26 & 3.37 & 36.12 & 28.73 \\
 & FLIP & 98.50 & 0.43 & 86.11 & 4.12 & 54.02 & 5.32 & 95.32 & 5.02 & 36.07 & 17.09 \\
 \rowcolor{gray!15}{\cellcolor{white}}& FLDetector & 98.54 & 0.39 & 84.30 & 3.62 & 54.08 & 1.12 & 95.13 & 0.19 & 33.41 & 29.55\\
 & FedREDefense & \textbf{98.64} & 0.33 & 85.52 & 2.37 & 53.33 & 0.55 & 95.26 & \textbf{0.12} & 35.03 & 0.82 \\
 \rowcolor{gray!45}{\cellcolor{white}\multirow{-12}{*}{$\alpha = 1.0$}}& FilterFL (Ours) & \textbf{98.64} & \textbf{0.22} & \textbf{86.70} & 2.27 & \textbf{54.56} & 0.59 & 95.57 & 0.14 & \textbf{36.92} & \textbf{0.63} \\ \midrule
 & No Attack & 98.38 & 0.22 & 85.52 & 2.51 & 54.31 & 0.50 & 95.63 & 0.20 & 36.45 & 0.41 \\
 & No Defense & 98.46 & 99.83 & 84.92 & 92.80 & 56.36 & 99.52 & 95.51 & 99.60 & 36.48 & 99.89 \\ \cmidrule(l){2-12} 
 \rowcolor{gray!15}{\cellcolor{white}}& Krum & 93.05 & 1.23 & 70.66 & 7.22 & 46.27 & 0.64 & 94.30 & 0.22 & 24.12 & 0.57 \\
 & COMED & 98.14 & 0.62 & 84.08 & 61.30 & 51.81 & 99.43 & 95.46 & 91.59 & 30.87 & 99.68 \\
 \rowcolor{gray!15}{\cellcolor{white}}& RLR & 93.85 & 1.84 & 76.93 & 8.30 & 26.64 & 42.25 & 94.13 & 34.26 & 20.73 & 68.87 \\
 & Trimmed Mean & 98.19 & 0.69 & 84.59 & 64.54 & 52.04 & 99.47 & \textbf{95.70} & 91.59 & 31.07 & 99.87 \\
 \rowcolor{gray!15}{\cellcolor{white}}& DP & 90.61 & 85.66 & 71.23 & 84.21 & 41.56 & 89.75 & 90.44 & 88.80 & 26.12 & 81.19 \\
 & FLAME & 98.16 & 0.37 & 81.85 & 93.78 & 51.52 & 98.24 & 94.76 & 66.78 & 34.96 & 89.15 \\
 \rowcolor{gray!15}{\cellcolor{white}}& FLTrust & 98.35 & 0.31 & 82.55 & 20.53 & 53.45 & 29.69 & 95.45 & 25.03 & 35.28 & 33.67 \\
 & FLIP & 98.43 & 1.32 & 85.29 & 10.23 & 54.32 & 22.32 & 95.36 & 12.56 & 35.12 & 28.71 \\
 \rowcolor{gray!15}{\cellcolor{white}}& FLDetector & 97.51 & 0.40 & 82.25 & 10.17 & 51.98 & 7.61 & 94.49 & 6.63 & 30.27 & 86.71 \\
 & FedREDefense & 98.52 & 0.38 & 81.32 & 2.50 & 51.46 & \textbf{0.61} & 94.28 & 10.21 & 34.15 & 12.03\\
 \rowcolor{gray!45}{\cellcolor{white}\multirow{-12}{*}{$\alpha=0.5$}}& FilterFL (Ours) & \textbf{98.55} & \textbf{0.31} & \textbf{85.46} & \textbf{2.48} & \textbf{54.95} & 0.65 & 95.45 & \textbf{0.14} & \textbf{35.32} & \textbf{0.57} \\ \midrule
 & No Attack & 97.65 & 0.32 & 74.17 & 6.50 & 50.97 & 0.58 & 96.36 & 0.06 & 32.16 & 0.57 \\
 & No Defense & 97.53 & 99.75 & 77.12 & 97.04 & 53.24 & 99.41 & 96.24 & 98.79 & 33.74 & 99.47 \\ \cmidrule(l){2-12} 
 \rowcolor{gray!15}{\cellcolor{white}}& Krum & 10.60 & 10.00 & 39.08 & \textbf{3.74} & 25.93 & 0.82 & 44.29 & 16.46 & 12.30 & 25.07 \\
 & COMED & 95.11 & 0.71 & 64.37 & 98.59 & 36.51 & 99.63 & 93.32 & 92.09 & 28.46 & 99.89 \\
 \rowcolor{gray!15}{\cellcolor{white}}& RLR & \textbf{97.32} & 98.88 & 68.66 & 89.36 & 19.80 & 57.12 & 95.92 & 99.08 & 27.98 & 99.22 \\
 & Trimmed Mean & 95.97 & 1.79 & 66.81 & 97.08 & 38.19 & 99.74 & 95.83 & 98.69 & 28.39 & 99.54 \\
 \rowcolor{gray!15}{\cellcolor{white}}& DP & 88.95 & 89.46 & 56.73 & 89.35 & 37.44 & 89.73 & 89.93 & 86.79 & 20.03 & 79.55 \\
 & FLAME & 94.43 & 90.17 & 66.89 & 98.86 & 46.68 & 99.61 & 93.92 & 97.41 & 29.87 & 95.42 \\
 \rowcolor{gray!15}{\cellcolor{white}}& FLTrust & 97.12 & 61.12 & 69.63 & 65.72 & 45.37 & 36.12 & 94.78 & 72.55 & 31.02 & 31.05 \\
 & FLIP & 96.63 & 23.21 & 72.71 & 24.52 & 47.96 & 30.33 & 95.07 & 35.12 & 30.78 & 29.32 \\
 \rowcolor{gray!15}{\cellcolor{white}}& FLDetector & 96.12 & 0.96 & 69.86 & 80.57 & 43.98 & 12.19 & 94.18 & 11.17 & 30.33 & 89.26\\
 & FedREDefense & 97.31 & 0.81 & 65.56 & 7.42 & 45.41 & 0.88 & 95.89 & 36.82 & \textbf{31.43} & 22.17\\
 \rowcolor{gray!45}{\cellcolor{white}\multirow{-12}{*}{$\alpha=0.1$}}& FilterFL (Ours) & 97.23 & \textbf{0.70} & \textbf{72.63} & 7.42 & \textbf{50.70} & \textbf{0.60} & \textbf{96.04} & \textbf{0.48} & 31.20 & \textbf{0.83} \\ \bottomrule
\end{tabular}
}
\end{table}

\end{document}